\documentclass[10pt,twocolumn,letterpaper]{article}

\usepackage[pagenumbers]{cvpr} % To force page numbers, e.g. for an arXiv version

\usepackage[dvipsnames]{xcolor}

\usepackage{amsmath,amsfonts,bm}

\def\eqref#1{equation~\ref{#1}}
\def\1{\bm{1}}

\def\vtheta{{\bm{\theta}}}

\def\vf{{\bm{f}}}

\def\vk{{\bm{k}}}
\def\vl{{\bm{l}}}
\def\vm{{\bm{m}}}

\def\vw{{\bm{w}}}
\def\vx{{\bm{x}}}
\def\vy{{\bm{y}}}

\DeclareMathAlphabet{\mathsfit}{\encodingdefault}{\sfdefault}{m}{sl}
\SetMathAlphabet{\mathsfit}{bold}{\encodingdefault}{\sfdefault}{bx}{n}

\definecolor{cvprblue}{rgb}{0.21,0.49,0.74}
\usepackage[pagebackref,breaklinks,colorlinks,citecolor=cvprblue]{hyperref}

\usepackage{multirow}
\usepackage{float}
\usepackage{indentfirst}
\usepackage{algorithm}
\usepackage[noend]{algorithmic}

\title{Bayesian Exploration of Pre-trained Models for Low-shot Image Classification}

\author{Yibo Miao$^{1}$, Yu Lei$^{1}$, Feng Zhou$^{2}$\footnotemark[1]\,, Zhijie Deng$^{1}$\thanks{Corresponding authors.} \\
        \textsuperscript{1}Qing Yuan Research Institute, SEIEE, Shanghai Jiao Tong University \\
        \textsuperscript{2}Center for Applied Statistics and School of Statistics, Renmin University of China\\
        \{miaoyibo, tony-lei, zhijied\}@sjtu.edu.cn, feng.zhou@ruc.edu.cn \\
}

\begin{document}
\maketitle
\begin{abstract}
Low-shot image classification is a fundamental task in computer vision, and the emergence of large-scale vision-language models such as CLIP has greatly advanced the forefront of research in this field. 
However, most existing CLIP-based methods lack the flexibility to effectively incorporate other pre-trained models that encompass knowledge distinct from CLIP.
To bridge the gap, this work proposes a simple and effective probabilistic model ensemble framework based on Gaussian processes, which have previously demonstrated remarkable efficacy in processing small data. 
We achieve the integration of prior knowledge by specifying the mean function with CLIP and the kernel function with an ensemble of deep kernels built upon various pre-trained models. 
By regressing the classification label directly, our framework enables analytical inference, straightforward uncertainty quantification, and principled hyper-parameter tuning. 
Through extensive experiments on standard benchmarks, we demonstrate that our method consistently outperforms competitive ensemble baselines regarding predictive performance. 
Additionally, we assess the robustness of our method and the quality of the yielded uncertainty estimates on out-of-distribution datasets. 
We also illustrate that our method, despite relying on label regression, still enjoys superior model calibration compared to most deterministic baselines.
\end{abstract}    
\section{Introduction}
\label{sec:intro}

The past few years have witnessed the trend of training large-scale foundation models to serve as infrastructures for processing images, texts, and multi-modal data~\cite{chen2020simple,he2022masked,devlin2018bert,brown2020language,radford2021learning,jia2021scaling}. 
The increasing availability of off-the-shelf pre-trained models is changing the standard practice for solving specific downstream tasks for AI practitioners. 
One fundamental application in vision is adapting pre-trained models for low-shot image classification. 
This eliminates the need for massive labeled data as in traditional cases, helps initiate the data annotation process, and supports the construction of complex recognition systems, among other advantages. 

Fine-tuning and linear probing are typical approaches for pre-trained models-based low-shot image classification~\cite{kolesnikov2020big,chen2020simple,chen2020big,caron2021emerging}. 
Recently, vision-language models, e.g., CLIP~\cite{radford2021learning}, have significantly advanced zero-shot classification where the image and semantics of interest are projected into a structured hidden space for nearest neighbor-based classification. 
Nevertheless, the few-shot CLIP with linear probing shows inferior results~\cite{radford2021learning}. 
To address this, researchers have put considerable effort into developing novel CLIP-based few-shot learning pipelines involving techniques such as prompting learning~\cite{zhou2022learning}, image-guided prompt generation~\cite{zhou2022conditional,qiu2021vt}, adapter tuning~\cite{gao2021clip,zhang2021tip}, etc.
Despite relatively good results, existing CLIP-based methods usually lose the flexibility to incorporate other pre-trained models that may contain complementary prior information. 

CaFo~\cite{zhang2023prompt} is a seminal work that explores constructing few-shot predictors using pre-trained models other than CLIP and demonstrates outperforming effectiveness. 
However, the ensemble weights in CaFo are determined heuristically, and the learning requires extensive hyper-parameter tuning. 
Furthermore, as a deterministic method, CaFo is likely to overfit the few-shot training data and cannot provide accurate uncertainty estimates. 
These challenges are particularly troublesome in situations with limited data and high-risk domains.

This paper aims to assemble CLIP and other pre-trained models in a more principled probabilistic manner.
Given that previous studies usually deploy a linear classification head on top of the pre-trained models, we focus on its Bayesian counterpart, i.e., a \emph{Gaussian process} (GP)~\cite{williams2006gaussian}. 
GP is an ideal model for low-shot image classification due to 
its effectiveness with \emph{small} data. 
To incorporate prior knowledge from various pre-trained models, we suggest defining the prior kernel as a combination of deep kernels associated with various pre-trained models.
Noting that the prior mean implicitly corresponds to a model that makes predictions without seeing any data, we specify it with the well-performing zero-shot CLIP classifier.

Such a modeling can address overfitting 
and result in calibrated \emph{post-data} uncertainty arising from posterior inference. 
Further, the Bayesian framework allows for the use of principled objectives for hyper-parameter tuning. 
For example, we can use the marginal likelihood or predictive likelihood of the GP model for hyper-parameter tuning following common practice.

We begin by assessing the predictive performance of our proposed method on standard low-shot image classification benchmarks and observe superior or competitive results compared to a variety of ensemble baselines. To evaluate the generalization capability of our method, we test the trained models on natural out-of-distribution (OOD) data and find that our method achieves outperforming results. In addition, our method has the potential to yield calibrated uncertainty estimates for OOD data. 
We further assess the model calibration by inspecting Expected Calibration Error (ECE)~\cite{guo2017calibration} and its more robust variant, Thresholded Adaptive Calibration Error (TACE)~\cite{nixon2019measuring}.
We also offer thorough ablation studies to better understand the proposed method.

\section{Related Works}
\textbf{Zero/few-shot classification. }
Few-shot classification means making classifications based on a limited number of observations, and the zero-shot one requires the trained model to adapt to the new task without any observation.
Meta-learning has demonstrated its potential as a viable approach for zero/few-shot learning~\cite{thrun1998learning,schmidhuber1987evolutionary}.
Recently, benefiting from the learning on web-scale data, large pre-trained vision-language models like CLIP have demonstrated impressive performance in zero/few-shot image classification. Since then, continual effort has been made to better adapt CLIP to downstream few-shot tasks~\cite{gao2021clip,zhang2021tip,zhou2022learning,zhou2022conditional,guo2022calip,udandarao2022sus,zhang2023prompt}. 
In particular, CoOp~\cite{zhou2022learning} optimizes a collection of learnable prompt tokens for few-shot adaptation. 
Tip-Adapter~\cite{zhang2021tip} augments the zero-shot CLIP classifier with a linear key-cache cache model to further enhance the classification performance. 
CaFo~\cite{zhang2023prompt} supplements Tip-Adapter with one further linear key-cache cache model for knowledge integration. 
Although effective, the deterministic nature of these methods makes them tend to overfit the few-shot training data and struggle to estimate predictive uncertainty. 

\textbf{Pre-trained models in vision and beyond. }
We have witnessed the change of model architectures in vision from VGG~\cite{simonyan2014very} and ResNet~\cite{he2016deep} to ViT~\cite{dosovitskiy2020image} and Swin Transformer~\cite{liu2021swin}.
The dominant learning paradigm has undergone a transformation, where pre-training models on extensive datasets and then utilizing them for downstream tasks via fine-tuning~\cite{he2019rethinking} has become a widespread practice. 
MoCo~\cite{chen2021empirical} and DINO~\cite{caron2021emerging} are recent representative pre-trained models, enjoying the ability to generate high-quality representations. 
Visual pre-trained models are instrumental in achieving state-of-the-art performance on diverse downstream tasks such as object detection~\cite{lin2017focal}, semantic segmentation~\cite{chen2017deeplab}, and so on. 
Recently, visual-language pre-training has achieved impressive success by learning from massive image-text pairs gathered from the internet~\cite{radford2021learning,jia2021scaling,li2023blip}, demonstrating astonishing performance on various downstream vision and language tasks.
We have reached the consensus that pre-trained models can serve as containers of valuable prior knowledge, but a proper mechanism for effective knowledge integration is under-explored, which is alleviated by this work. 

\textbf{Deep Gaussian processes. }
GPs are a well-studied and powerful probabilistic tool in machine learning~\cite{williams2006gaussian}. They share a deep connection with neural networks (NNs) with infinite width~\cite{neal1996bayesian,lee2017deep,jacot2018neural}.
There exists an interesting correspondence among linear regression, Bayesian linear regression, and GP regression, with the last one often preferred in low-data regimes. 
GPs have been successfully used to solve classification problems based on approximations~\cite{nickisch2008approximations}.
However, GPs built on classic kernels lack the inductive bias carried by NNs. 
To address this, deep kernel learning (DKL)~\cite{calandra2016manifold,wilson2016deep} has been proposed to leverage deep NNs for nonlinear data projection, which is then fed to classic kernels. 
In the context of few-shot learning, deterministic methods with a linear classification head face challenges of overfitting to the training set and are unable to accurately quantify uncertainty. This limitation restricts their applicability in high-risk domains. In contrast, GPs offer a viable remedy to these pathologies. 

\section{Preliminary}
This section reviews the basics of GP regression and deep kernel learning. 
We use $\mathcal{D} = \{\vx_i,\vy_i\}_{i=1}^N$ to denote a dataset with $\vx_i \in\mathcal{X}\subset\mathbb{R}^L$ and $\vy_i \in\mathbb{R}^C$ as the $L$-dim inputs and $C$-dim targets respectively.
Let $\mathbf{X}=\{\vx_i\}_{i=1}^N$ and $\mathbf{Y}=\{\vy_i\}_{i=1}^N$ represent the training data. 
Let $\mathbf{X}^{\text{val}}$ and $\mathbf{Y}^{\text{val}}$ represent the validation data (can be split from the training data) 
and $\mathbf{X}^*=\{\vx_i^{*}\}_{i=1}^M$ represent the test data.

\subsection{From Deterministic to Bayesian}

To deal with the learning problem on the above dataset, it is common practice to train a deterministic model $f: \mathcal{X} \to \mathbb{R}^C$ using maximum likelihood estimation or maximum a posteriori principle. 
Despite effectiveness, the approach can suffer from detrimental overfitting and struggle to reason about model uncertainty appropriately. 
These issues are exacerbated when only limited data is available. 

Practitioners can turn to Bayesian learning approaches to address such issues. In Bayesian learning, a prior distribution over model parameters 
is introduced, and the Bayesian posterior is (approximately) computed. 
Then, we compute the posterior predictive distribution to predict for a new datum, where all likely model specifications are considered.  
The uncertainty can be quantified by certain statistics that capture the degree of variation in that distribution. 

\subsection{Gaussian Process Regression}

GP regression is an extensively studied function-space Bayesian model~\cite{williams1995gaussian}. It enjoys exact Bayesian inference and non-parametric flexibility, allowing for a high degree of freedom in kernel specification to adapt the model to various types of nonlinear data. Consequently, it is often the preferred choice for \emph{small-} to \emph{medium-}sized datasets.

Specifically, GP regression usually deploys a prior in the following formula: 
\begin{equation}
    f(\vx) \sim \mathcal{GP}(m(\vx),k(\vx,\vx')),
\end{equation}
where $m(\vx)$ indicates the mean function and $k(\vx,\vx')$ denotes the kernel (covariance) function that describes the similarity among data points. 
Assume additive isotropic Gaussian noise on the function output, which corresponds to a Gaussian likelihood $y(\vx) | f(\vx) \sim \mathcal{N}(y(\vx); f(\vx), \sigma^2\mathbf{I})$ where $\sigma^2$ is the noise variance. 
The predictive distribution of the function evaluations $\vf^*$ on new data points $\mathbf{X}^*$ is: 
\begin{equation}
\begin{aligned}
& \vf^* | \mathbf{X}^*, \mathbf{X}, \mathbf{Y} \sim \mathcal{N}(\mathbb{E}[\vf^*], \operatorname{cov}(\vf^*)),
\end{aligned}
\end{equation}
where
\begin{equation}
\small
\begin{aligned}
\label{eq:3}
\mathbb{E}[\vf^*] &:=\vm_{\mathbf{X}^*}+\vk_{\mathbf{X}^*, \mathbf{X}}[\vk_{\mathbf{X}, \mathbf{X}}+\sigma^2 \mathbf{I}]^{-1}(\mathbf{Y}-\vm_\mathbf{X}), \\
\operatorname{cov}(\vf^*) &:=\vk_{\mathbf{X}^*, \mathbf{X}^*}-\vk_{\mathbf{X}^*, \mathbf{X}}[\vk_{\mathbf{X}, \mathbf{X}}+\sigma^2 \mathbf{I}]^{-1} \vk_{\mathbf{X}, \mathbf{X}^*}, 
\end{aligned}
\end{equation}
$\vm_\mathbf{X} \in \mathbb{R}^{N \times C}$ and $\vk_{\mathbf{X}, \mathbf{X}} \in \mathbb{R}^{N \times N}$ represent the evaluation of $m(\cdot)$ and $k(\cdot, \cdot)$ on the training data $\mathbf{X}$ respectively. 
Other matrices are defined similarly. 
Unlike parametric models such as NNs, GP makes predictions for new data by referring to the training samples, similar to how humans approach the task. 

This model can be readily adapted to tackle classification problems by treating the one-hot labels as regression targets, 
which is known as the label regression~\cite{kuss2006gaussian,patacchiola2020bayesian,lee2017deep}. The label regression design enables analytical expressions for both evidence and posterior, making the classifier computationally efficient and easy to implement.

The GP regression also offers analytical objectives for tuning parameters (denoted as $\boldsymbol{\alpha}$; including $\sigma^2$ and others in the definition of $m$ and $k$). 
One typical choice is the log marginal likelihood:
\begin{equation}
    \label{eq:marginal}
    \begin{split}
    & \log p(\mathbf{Y}|\mathbf{X}, \boldsymbol{\alpha}) \propto -[\operatorname{trace}((\mathbf{Y}-\vm_\mathbf{X})^\top (\vk_{\mathbf{X}, \mathbf{X}} \\
    & \quad\quad+ \sigma^2 \mathbf{I})^{-1} (\mathbf{Y}-\vm_\mathbf{X})) + C\log |\vk_{\mathbf{X}, \mathbf{X}} + \sigma^2 \mathbf{I}|],
    \end{split}
\end{equation}
which corresponds to the summation of the log marginal likelihood of $C$ independent $1$-dim GP regressions. 
Yet, it is shown that this objective can be negatively correlated with the generalization~\cite{lotfi2022bayesian,ke2023revisiting}. 
Given this, a more proper objective can be $\log p(\mathbf{Y}^\text{val}|\mathbf{X}^\text{val}, \mathbf{X}, \mathbf{Y}, \boldsymbol{\alpha})$, i.e., the predictive likelihood on extra validation data $(\mathbf{X}^\text{val}, \mathbf{Y}^\text{val})$. 
It also takes the form of Gaussian log densities.

\subsection{Deep Kernel Learning}

In DKL, a $\vtheta$-parameterized deep NN $g_\vtheta: \mathcal{X} \to \mathbb{R}^D$ is typically used to transform the input data $\vx$ into hidden features $g_\vtheta(\vx)$. The kernel is then defined as:
\begin{equation}
k(\vx,\vx') := \tilde{k}(g_\vtheta(\vx), g_\vtheta(\vx')),
\end{equation}
where $\tilde{k}$ is a base kernel, such as the popular radial basis function (RBF) kernel or polynomial kernel.

To make the NN parameters better suited for the data at hand, DKL treats them as hyper-parameters of the GP model and optimizes them to maximize the marginal likelihood. However, the large number of hyper-parameters makes the optimization time-consuming and increases the risk of overfitting~\cite{ober2021promises}. It can even underperform a standard deterministic NN in some toy cases. 
\section{Methodology}

\begin{figure*}[t]
    \centering
    \includegraphics[width={\linewidth}]{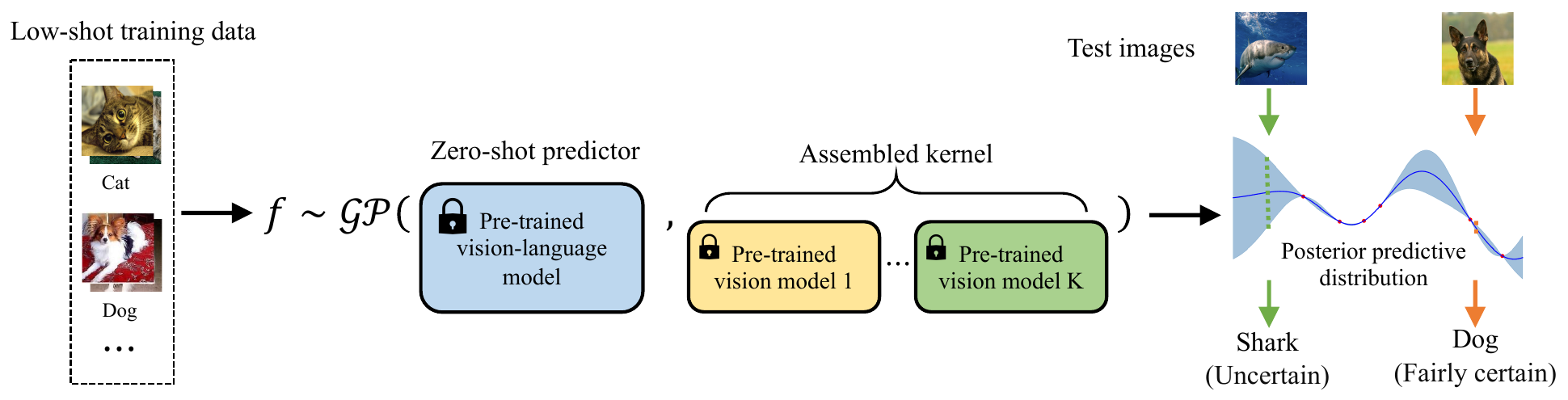}
    \caption{
    Overview of our method. We leverage a GP regressor to tackle the low-shot image classification problem. 
    To integrate knowledge from CLIP and other pre-trained models, we use them to specify the GP mean and kernel. 
    The label is determined by the mean, and the uncertainty estimate is determined by the variance.
    }
    \vspace{-2pt}
    \label{fig:framework}
\end{figure*}

This section explores a Bayesian approach to assemble CLIP with other pre-trained models for low-shot image classification. 
Given the discussion above, we take the GP regression as the modeling framework. 
We then elaborate on how to integrate various pre-trained models into it. 
We provide an overview of our method in \cref{fig:framework}.

\subsection{Design of Kernel}
Utilizing an NN-based feature extractor to define the kernel function aids to incorporate informative inductive bias into GP, which is essential for processing complex data such as images and texts. 
However, conventional approaches like DKL suffer from the pathology that all involved NN parameters are required to be carefully tuned. 
Considering that pre-trained models can yield representations that are generally applicable to a wide range of applications, 
we propose to alternatively use pre-trained models to define deep kernels and then perform an adaptive combination. 
By doing this, the number of hyper-parameters in the GP is reduced significantly, and the prior knowledge encoded by various pre-trained models is effectively integrated.

Specifically, assuming access to $K$ pre-trained models $g^{(i)}: \mathcal{X} \to \mathbb{R}^{D^{(i)}}, i=1,\ldots,K$,\footnote{We omit the dependency of these models on their parameters because we regard them as fixed models and do not perform fine-tuning. We assume an $L^2$ normalization at the end of each model unless specified otherwise.} 
we define the following independent kernels: 
\begin{equation}
    \begin{split}
    & k^{(i)}(\vx, \vx') := \tilde{k}(\vl^{(i)} \circ g^{(i)}(\vx) , \vl^{(i)} \circ g^{(i)}(\vx') ),
    \end{split}
\end{equation}
where $\circ$ denotes the element-wise product and $\vl^{(i)} \in \mathbb{R}_+^{D^{(i)}}$ is a learnable vector used to boost flexibility, e.g., when the base kernel $\tilde{k}$ is the RBF kernel, $\vl^{(i)}$ defines the learnable length-scales for it. 
We can also enforce a constraint where all elements in $\vl^{(i)}$ have the same value, and the final learning outcomes are slightly impacted.
By summing up these kernels, we get the final kernel:
\begin{equation}
    k(\vx, \vx') := \sum_{i=1}^K  k^{(i)}(\vx, \vx').
\end{equation}
The learnable hyper-parameters enable an easy, automatic adaptation of the kernel to specific data.

\subsection{Design of Mean}
In essence, the prior mean $m(\cdot)$ refers to a function making predictions before seeing any data, i.e., a zero-shot predictor.
Traditionally, $m(\cdot)$ is set to zero for simplicity. 
However, as shown in \cref{sec:ablk}, this can lead to poor generalization performance in low-shot image classification tasks in practice. 
This suggests that it is necessary to incorporate effective prior knowledge of $m(\cdot)$ into the GP.

Interestingly, a similar phenomenon has been reported in the literature, where the linear probe CLIP using few-shot data performs much worse than zero-shot CLIP~\cite{radford2021learning}.
This is because the knowledge in the zero-shot CLIP classifier has not been effectively integrated into the few-shot learners.

With these insights, we make a simple yet significant improvement to our GP model. We set the mean function $m(\cdot)$ to the zero-shot linear classifier in CLIP, which has demonstrated strong performance. 
Concretely, 
let $g: \mathcal{X} \to \mathbb{R}^{D}$ denotes CLIP's image encoder, and $\vw \in \mathbb{R}^{D \times C}$ denotes the weight of the zero-shot linear classifier composed of embeddings of the text descriptions of the $C$ classes of interest. 
Our prior mean takes the following form:
\begin{equation}
    m(\vx) := \gamma \operatorname{softmax}(\tau g(\vx)^\top \vw),
\end{equation}
where $\tau, \gamma \in \mathbb{R}_+$ denote the introduced learnable temperature and scale respectively. 
Notably, we use a softmax operation to obtain classification probabilities directly because we formulate the classification problem as a regression one. 

\begin{algorithm}[t]
\caption{Leverage Gaussian processes to assemble pre-trained models for low-shot image classification}
\label{algo:1}
\begin{algorithmic}[1]
\STATE {\bfseries Input:} Number of optimization steps $T$, training data $\mathbf{X}$, $\mathbf{Y}$, validation data $\mathbf{X}^{\text{val}}$, $\mathbf{Y}^{\text{val}}$, test data $\mathbf{X}^*$, hyper-parameters $\boldsymbol{\alpha}$.
\STATE {\bfseries Output:} Predictions ($\mathbf{Y}^*$) of $\mathbf{X}^*$ and $\operatorname{cov}(\vf^*)$.
\FOR{$t=1\rightarrow T$ }
    \STATE{Obtain $\mathbb{E}[\vf^\text{val}]$ and $\operatorname{cov}(\vf^\text{val})$ of $\mathbf{X}^\text{val}$ via \cref{eq:3};}
    \STATE{ Calculate $\log p(\mathbf{Y}^\text{val}|\mathbf{X}^\text{val}, \mathbf{X}, \mathbf{Y}, \boldsymbol{\alpha})$ and estimate its gradients w.r.t. $\boldsymbol{\alpha}$;}
    \STATE{Update $\boldsymbol{\alpha}$ by one-step gradient ascent;}
\ENDFOR
\STATE{Obtain $\mathbb{E}[\vf^*]$ and $\operatorname{cov}(\vf^*)$ of $\mathbf{X}^*$ via \cref{eq:3};}
\STATE{$\mathbf{Y}^* = \operatorname{argmax}(\mathbb{E}[\vf^*])$;}
\end{algorithmic}
\end{algorithm}

\subsection{Learning}
Using the classification likelihood for data fitting is also viable. 
However, doing so naturally disrupts conjugacy, leading to the inability to estimate the posterior in a closed form~\citep{williams2006gaussian,bishop2006pattern}. 
Therefore, we advocate label regression for its computational efficiency and ease of implementation. It also allows us to revert to analytical expressions for both the evidence and the posterior.

\begin{table*}[!t]
    \setlength{\tabcolsep}{15pt}
    \centering
    \begin{tabular}{lccccccc}
        \toprule
        \multicolumn{1}{l}{Shot}  & 1 & 2 & 4 & 8 & 16   \\
        \midrule
         Ens-LP &	40.25 $\pm$ 0.09&	49.79 $\pm$ 0.07	&57.42 $\pm$ 0.06&	62.28 $\pm$ 0.09 &	66.31 $\pm$ 0.13\\
         Ens-LP$^\dagger$ &	61.77 $\pm$ 0.13&	64.10 $\pm$ 0.35 &65.89 $\pm$ 0.33 &67.59 $\pm$ 0.06 &69.83 $\pm$ 0.27\\
         Ens-CaFo &	62.09 $\pm$ 0.13 &63.67 $\pm$ 0.19 &64.96 $\pm$ 0.06 &	66.57 $\pm$ 0.42 &	68.78 $\pm$ 0.25\\
        Ours &\textbf{63.07 $\pm$ 0.07} & \textbf{65.17 $\pm$ 0.23} & \textbf{67.50 $\pm$ 0.06} & \textbf{69.31 $\pm$ 0.08} & \textbf{70.77 $\pm$ 0.07} \\
        \bottomrule
    \end{tabular}
    \caption{
    Comparison with ensemble baselines of low-shot classification accuracy (\%) on ImageNet.}
    \label{sample-table}
\end{table*} 

One extra merit of label regression is that it enables the tractable marginalization of data likelihood, so we can perform hyper-parameter tuning more easily. 
Let $\boldsymbol{\alpha}:=\{\sigma^2, \vl^{(1)}, \dots, \vl^{(K)},  \tau, \gamma\}$ denote all hyper-parameters in our method. 
We optimize them by maximizing the aforementioned log marginal likelihood or log predictive likelihood to make them suitable for the data. 
In the low-shot learning scenario, the dataset size is small, allowing us to compute the kernel matrix, its inversion, and its determinant with minimal cost. Using an Adam optimizer~\cite{kingma2014adam}, convergence is usually rapid, typically within $100$ optimization steps. We depict the overall algorithmic procedure in \cref{algo:1}.

\subsection{Uncertainty Quantification}
As per convention, we utilize the diagonal elements of $\operatorname{cov}(\vf^*)$ (outlined in \cref{eq:3}) to quantify the predictive uncertainty of our model on the test data points. 
This information enables us to refrain from making predictions on data with high uncertainty and, instead, implement other conservative fallback strategies to handle such situations. 
Moreover, we can leverage this information to identify OOD samples since they typically exhibit greater uncertainty than in-distribution data. 

\section{Experiments}

\begin{figure*}[t]
    \centering
    \includegraphics[width={\linewidth}]{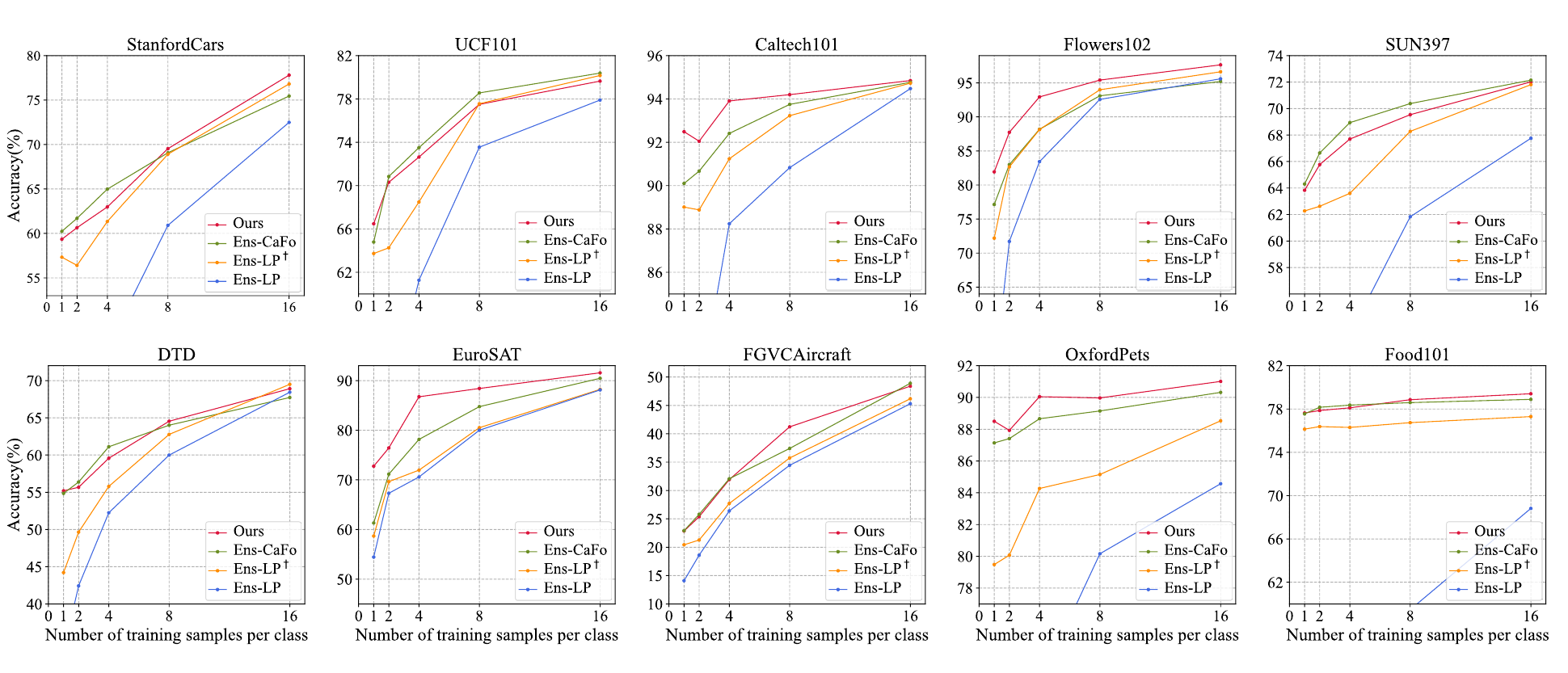}
    \caption{Comparison of low-shot classification accuracy (\%) on the ten popular benchmarks.
    }
    \vspace{-2pt}
    \label{fig:ten}
\end{figure*}

We first demonstrate that our method achieves competitive low-shot performance on diverse and prevalent benchmarks. Subsequently, we illustrate how our uncertainty estimates can identify OOD samples and validate the calibration of the learning outcomes. We also conduct ablation studies on our method and provide analyses.

\subsection{Experimental Setup}

\textbf{Datasets.} Following CaFo~\cite{zhang2023prompt}, we conduct experiments on image classification datasets including ImageNet~\cite{deng2009imagenet} and 10 other widely-used ones: StanfordCars~\cite{krause20133d}, UCF101~\cite{soomro2012ucf101}, Caltech101~\cite{fei2004learning}, Flowers102~\cite{nilsback2008automated}, SUN397~\cite{xiao2010sun}, DTD~\cite{cimpoi2014describing}, EuroSAT~\cite{helber2019eurosat}, FGVCAircraft~\cite{maji2013fine}, OxfordPets~\cite{parkhi2012cats}, and Food101~\cite{bossard2014food}.
We follow CaFo to train the model with 1, 2, 4, 8, and 16 shots of training data and test on the entire test set. 

\textbf{Pre-trained models.}
Unless specified otherwise, we use the ResNet-50 version of CLIP. 
Besides, we consider the model trained by MoCo with ResNet-50 architecture, and that trained by DINO with ResNet-50 architecture for ensemble due to their popularity. 
We clarify that other pre-trained models are readily applicable to our framework.

\textbf{Baselines.}
To validate that our ensemble strategy is non-trivial, we build three baselines for comparison: (1) Ens-LP, short for the ensemble of linear probing, where we apply linear probing to each pre-trained model and take the average of their output probabilities for prediction, 
(2) Ens-LP$^\dagger$, where the zero-shot CLIP  classifier is further integrated into Ens-LP, and (3) Ens-CaFo, where we generalize the original CaFo approach to assemble multiple pre-trained models by fusing logits. Notably, the
original CaFo approach uses images generated by DALL·E~\cite{ramesh2021zero} for data augmentation. 
We do not use this strategy for all results reported in our paper. 
\textbf{Training protocols.}
For the hyper-parameters, we initialize the noise variance $\sigma^2=0.01$, the scale $\gamma=1$, and the temperature $\tau=100$. 
The learnable length-scales $\vl$ of the kernel are randomly initialized. 
We can also constrain all elements in $\vl$ to have the same value, and the results are slightly impacted.\footnote{For example, on the 16-shot ImageNet, the accuracy is 70.77$\%$ when $\vl$ is set as a learnable vector and 70.42$\%$ when constraining all elements in $\vl$ to have the same value. }
Notably, since optimizing hyper-parameters using predictive likelihood requires a validation split, which is not feasible under 1-shot setting of ImageNet, we instead utilize marginal likelihood to optimize the hyper-parameters in that case. 
When using the predictive likelihood, there is an equal 1:1 ratio between the training and validation splits.
On other datasets, following CaFo~\cite{zhang2023prompt}, we tune the hyper-parameters by the official validation sets.
We perform hyper-parameter optimization for 100 steps with an Adam optimizer with a learning rate of $0.01$ (a cosine decay is adopted). 
The optimization is low-cost, e.g., only requiring about 4 minutes on a single RTX-3090 under the 16-shot ImageNet setting. 
We follow CaFo to construct the zero-shot CLIP classifier.
We report the average results over three random runs.

\subsection{Predictive Performance}

We first evaluate the low-shot classification performance on the ImageNet benchmark. 
The results are presented in \cref{sample-table}.
As shown, our method outperforms other baselines with clear margins. 
The less favorable results of the baselines underscore the inherent challenges in amalgamating knowledge from multiple pre-trained models. 

The merits of our method are more prominent for medium-sized training data (e.g., the 4 and 8 shots).
It is worth noting that the performance difference between Ens-LP and Ens-LP$^\dagger$ is quite significant,
which further underscores the importance of introducing the zero-shot CLIP-based classifier.

To further evidence the generality and superiority of our model, we conduct experiments on ten other popular benchmarks across various domains, with the results reported in \cref{fig:ten}.
As shown, our method surpasses or is on par with the competing baselines on most benchmarks. 

\begin{table}[t]
    \setlength{\tabcolsep}{5pt}
    \centering
    \begin{tabular}{lccccccc}
        \toprule
        \multicolumn{1}{l}{Shot}  & 1 & 2 & 4 & 8 & 16             \\
        \midrule
        Linear-probe  & 22.17 & 31.90 & 41.20 &49.52 & 56.13 \\
        CoOp  & 57.15 &57.81 & 59.99 &61.56 & 62.95 \\
        CLIP-Adapter  & 61.20 & 61.52 & 61.84 & 62.68 & 63.59 \\
        VT-CLIP  & 60.53 &61.29 & 62.02 & 62.81 & 63.92 \\
        Tip-Adapter-F  &61.32 & 61.69 & 62.52 & 64.00 &65.51 \\
        CALIP-FS  &61.35 & 62.03 & 63.13 & 64.11 &65.81 \\
        Ours & \textbf{63.07} & \textbf{65.17} & \textbf{67.50 } & \textbf{69.31 } & \textbf{70.77 }\\
        \bottomrule
    \end{tabular}
    \caption{
    Comparison with leading methods of low-shot classification accuracy (\%) on ImageNet.}
    \label{tab:leading}
\end{table}

We also compare our method to leading CLIP-based low-shot learners, including 
CLIP-Adapter~\cite{gao2021clip}, Tip-Adapter-F~\cite{zhang2021tip}, CoOp~\cite{zhou2022learning}, and CALIP-FS~\cite{guo2022calip} on ImageNet~\cite{deng2009imagenet}. All these methods use the CLIP model with ResNet-50 architecture, the same as ours.
The results in \cref{tab:leading}
show that our method consistently achieves higher accuracy than the leading approaches,  
which indicates the necessity of assembling complementary prior knowledge from various pre-trained models for low-shot classification.

\begin{figure*}[t]
  \begin{subfigure}{0.25\linewidth}
    \centering
    \includegraphics[width=\linewidth]{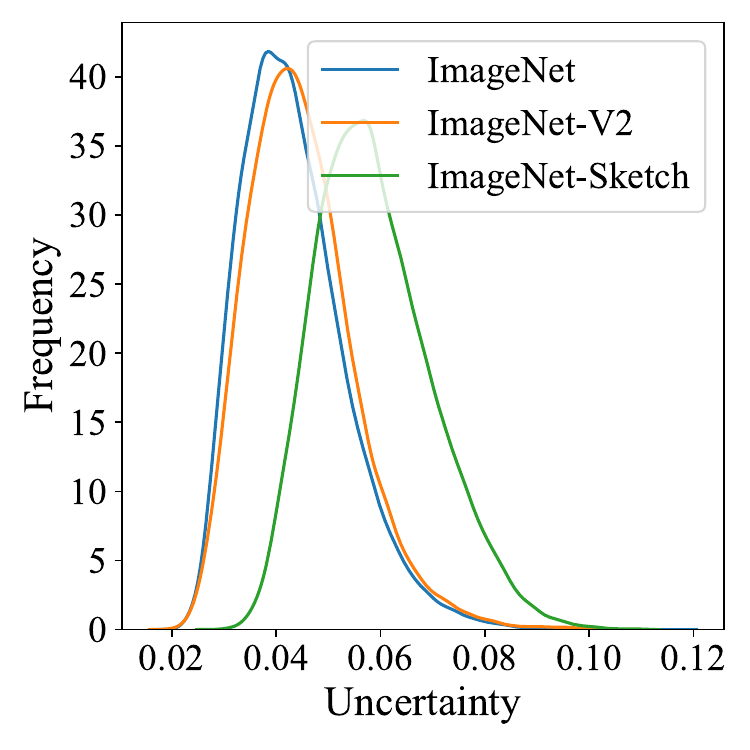}
    \caption{Ours}
    \label{ours_unc}
  \end{subfigure}%
  \begin{subfigure}{0.25\linewidth}
    \centering
    \includegraphics[width=\linewidth]{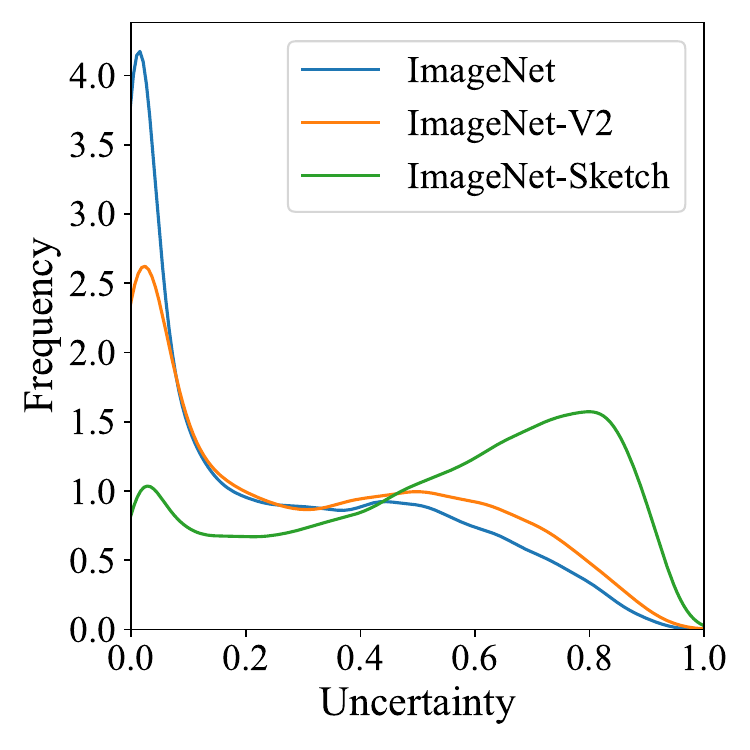}
    \caption{Ens-CaFo.}
  \end{subfigure}%
  \begin{subfigure}{0.25\linewidth}
    \centering
    \includegraphics[width=\linewidth]{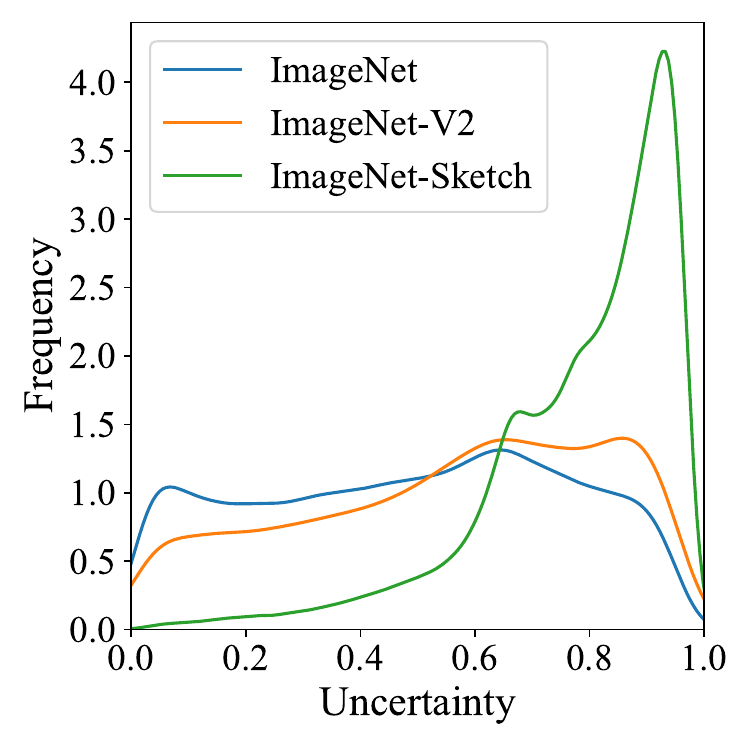}
    \caption{Ens-LP.}
  \end{subfigure}%
  \begin{subfigure}{0.25\linewidth}
    \centering
    \includegraphics[width=\linewidth]{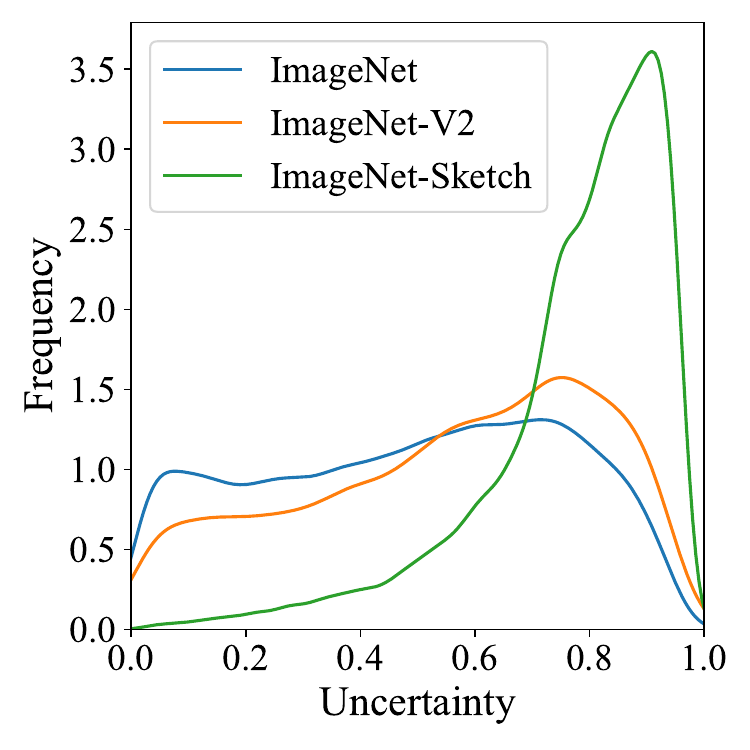}
    \caption{Ens-LP$^\dagger$.}
  \end{subfigure}%
  \caption{Histogram for uncertainty estimates. We evaluate
different ensemble methods on ImageNet, ImageNet-V2, and Imagenet-Sketch.}
    \label{fig:ood_detection}
\end{figure*}

\subsection{Evaluation on OOD Data}
We next evaluate the robustness and the quality of uncertainty estimates of our method on OOD data.

\textbf{OOD robustness.}
We use our model trained on 16-shot ImageNet to evaluate OOD samples from ImageNet-V2~\cite{recht2019imagenet} and ImageNet-Sketch~\cite{hendrycks2021natural}.
ImageNet-v2 is an ImageNet test set collected using the original labeling protocol, with 10 samples per class. ImageNet-Sketch shares the same classes as ImageNet, but all images are sketches. 
As shown in \cref{OOD}, our model exhibits superior OOD robustness compared to the ensemble baselines on both ImageNet-V2 and ImageNet-Sketch.

 \begin{table}[t]
    \setlength{\tabcolsep}{12pt}
    \centering
    \begin{tabular}{lccc}
    \toprule
    \multirow{2} * {Datasets} & \textbf{Source} & \multicolumn{2}{c}{\textbf{Target}} \\
    \cmidrule(r){2-2}\cmidrule(r){3-4}
    & ImageNet & -V2 & -Sketch \\
    \midrule
    Ens-CaFo & 68.53 & 59.62 & 36.12 \\
    Ens-LP & 66.37 & 55.08 & 24.76 \\
    Ens-LP$^\dagger$ & 70.13 & 59.86 & 34.66 \\
    \textbf{Ours} & \textbf{70.77} & \textbf{61.30} & \textbf{36.58} \\
    \bottomrule
      \end{tabular}
     \captionof{table}{Test accuracy (\%) on OOD datasets.}
     \vspace{-2pt}
     \label{OOD}
  \end{table}

\begin{figure*}
  \begin{subfigure}{0.25\linewidth}
    \centering
    \includegraphics[width=\linewidth]{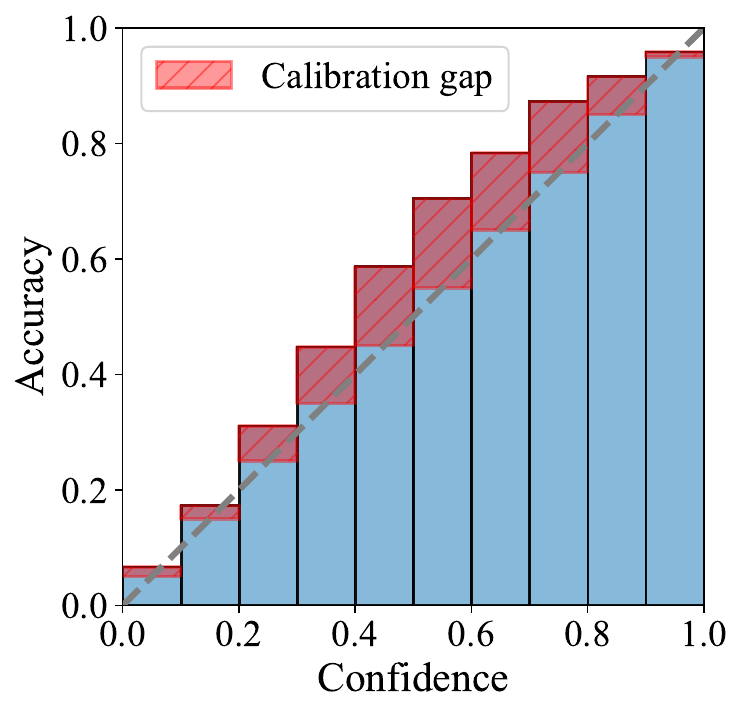}
    \caption{Ours}
  \end{subfigure}%
  \begin{subfigure}{0.25\linewidth}
    \centering
    \includegraphics[width=\linewidth]{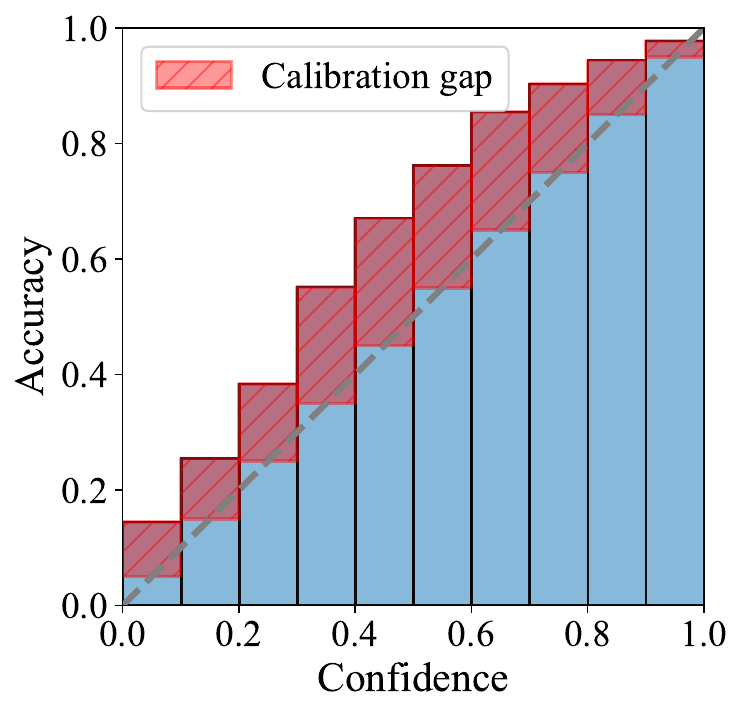}
    \caption{Ens-LP}
  \end{subfigure}%
  \begin{subfigure}{0.25\linewidth}
    \centering
    \includegraphics[width=\linewidth]{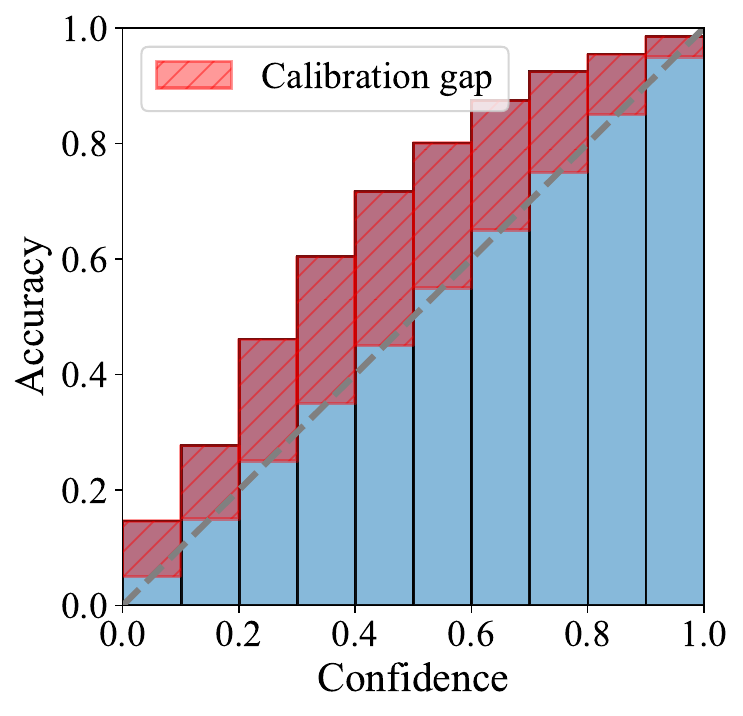}
    \caption{Ens-LP$^\dagger$}
  \end{subfigure}%
  \begin{subfigure}{0.25\linewidth}
    \centering
    \includegraphics[width=\linewidth]{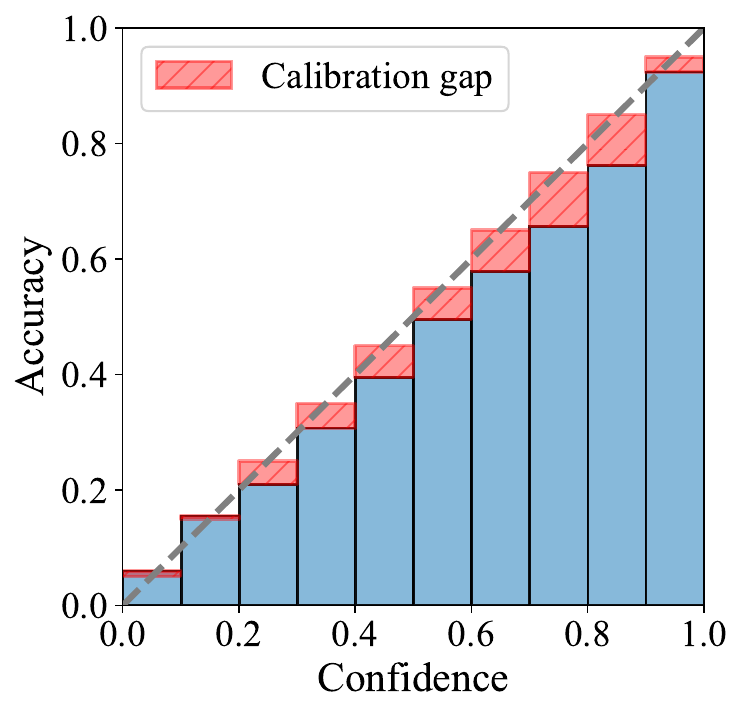}
    \caption{Ens-CaFo}
  \end{subfigure}%
  \caption{Realibility diagrams of the four ensemble methods.}
  \vspace{-4pt}
    \label{fig:ece}
\end{figure*}
\textbf{Quality of uncertainty estimates.} We then assess the quality of our uncertainty estimates on the above OOD datasets. 
We collect the predictive uncertainty estimates yielded by our model for both in-distribution data points and OOD ones and depict the histogram in \cref{fig:ood_detection}, where the results of the baselines are also included. 
As the baselines are deterministic, we take one minus the prediction confidence as their uncertainty estimate.
As implied by the histograms, our model does not regard ImageNet-V2 as OOD data, which aligns with the fact that the distribution of ImageNet-V2 is as similar as possible to the original ImageNet~\cite{qiu2021vt}. 
On the other hand, our method can clearly identify the OOD ImageNet-Sketch dataset. 
For the other three baselines, while we can also observe that the differences between ImageNet and ImageNet-V2 are smaller than those between ImageNet and ImageNet-Sketch, the manifestations of these properties are not as pronounced as in our approach. 
We also provide additional illustrative figures demonstrating the utilization of uncertainty estimates, shown in the Appendix. 

We further quantitatively estimate the effectiveness of the uncertainty estimates by using them to distinguish ImageNet-Sketch from ImageNet. 
The AUROCs for our method, Ens-LP, Ens-LP$^\dagger$, and Ens-CaFo to distinguish between ImageNet and ImageNet-Sketch are 0.8545, 0.8249, 0.8253, and 0.7546 respectively.
These results align with our previous analyses and validate the superior reliability of our uncertainty estimates.

\begin{figure*}[t]
 \centering
    \begin{subfigure}{0.25\linewidth}
      \includegraphics[width=\linewidth]{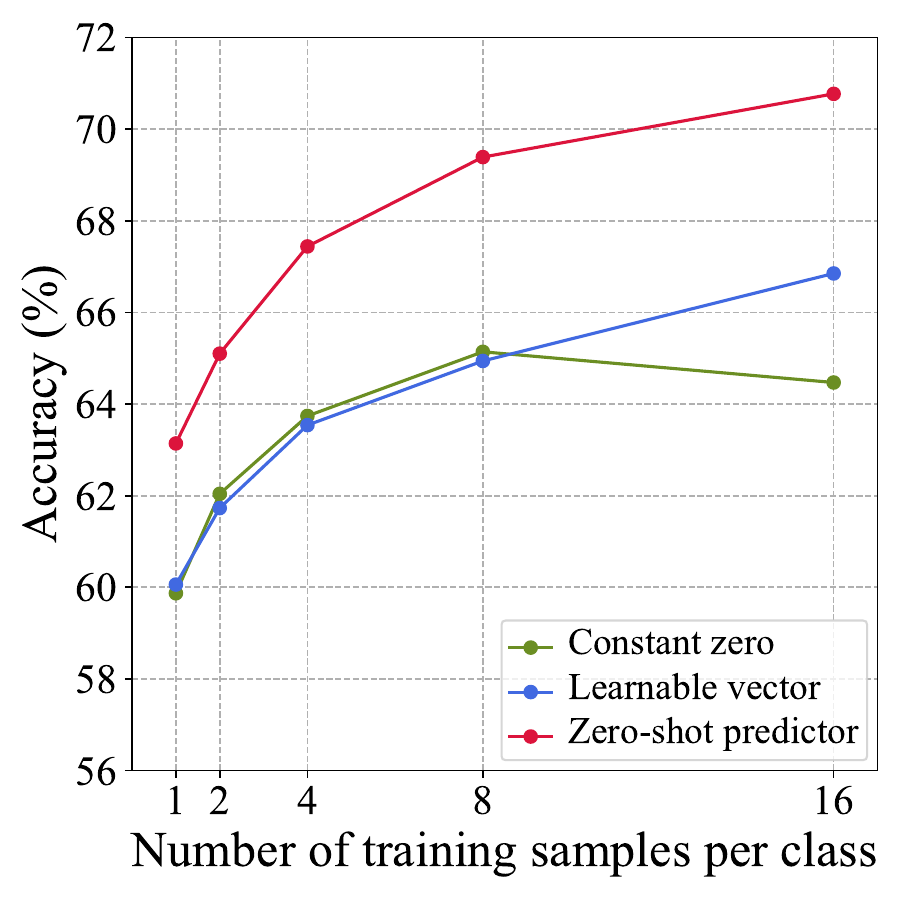}
      \caption{}
      \label{ablationmean}
    \end{subfigure}%
    \begin{subfigure}{0.25\linewidth}
      \includegraphics[width=\linewidth]{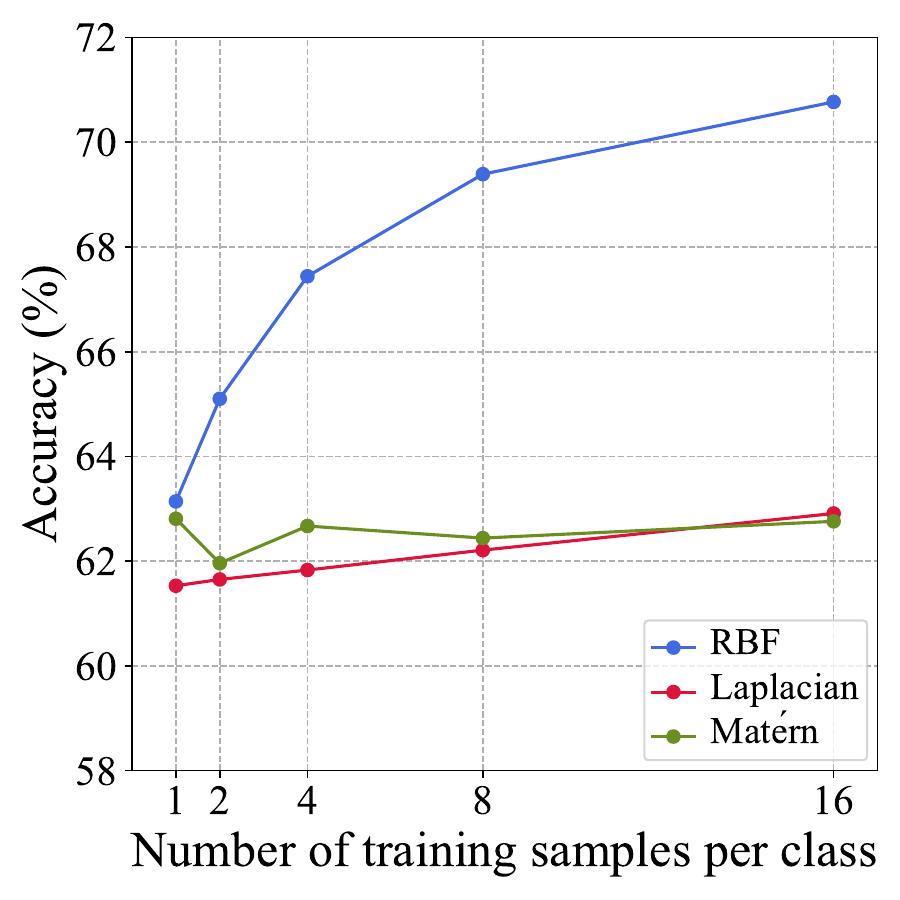}
      \caption{}
      \label{ablationcov} 
    \end{subfigure}%
    \begin{subfigure}{0.25\textwidth}
      \includegraphics[width=\linewidth]{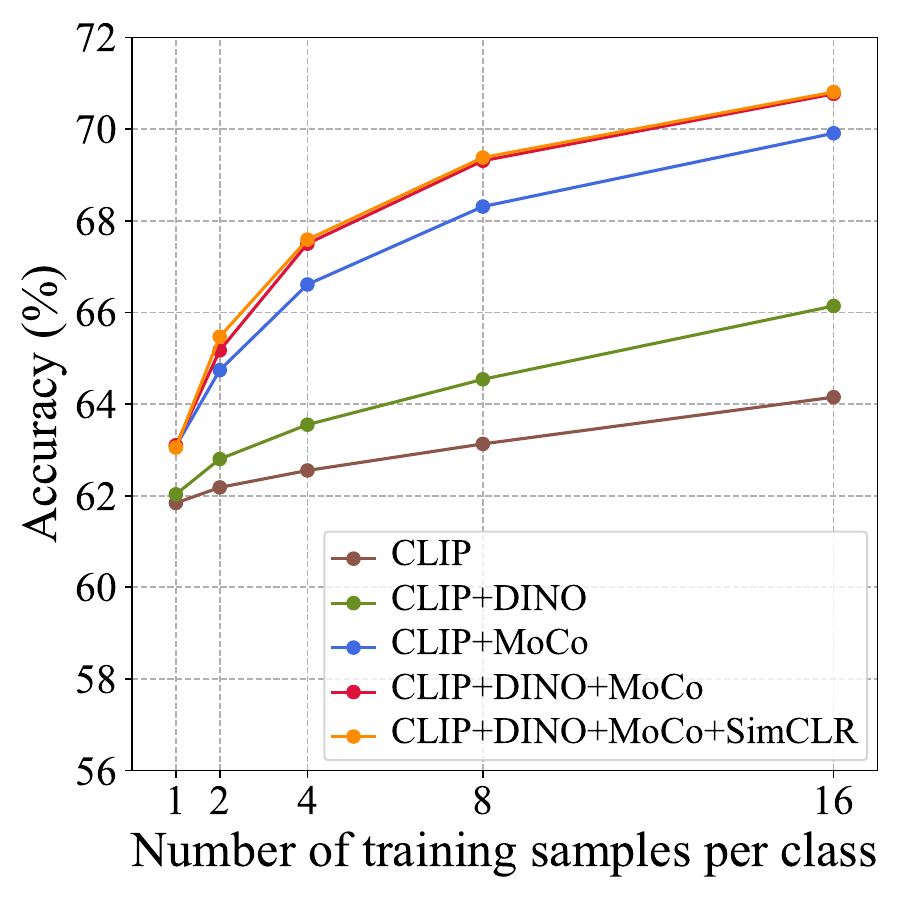}
      \caption{}
      \label{ablationkernel}
    \end{subfigure}%
    \begin{subfigure}{0.25\linewidth}
      \includegraphics[width=\linewidth]{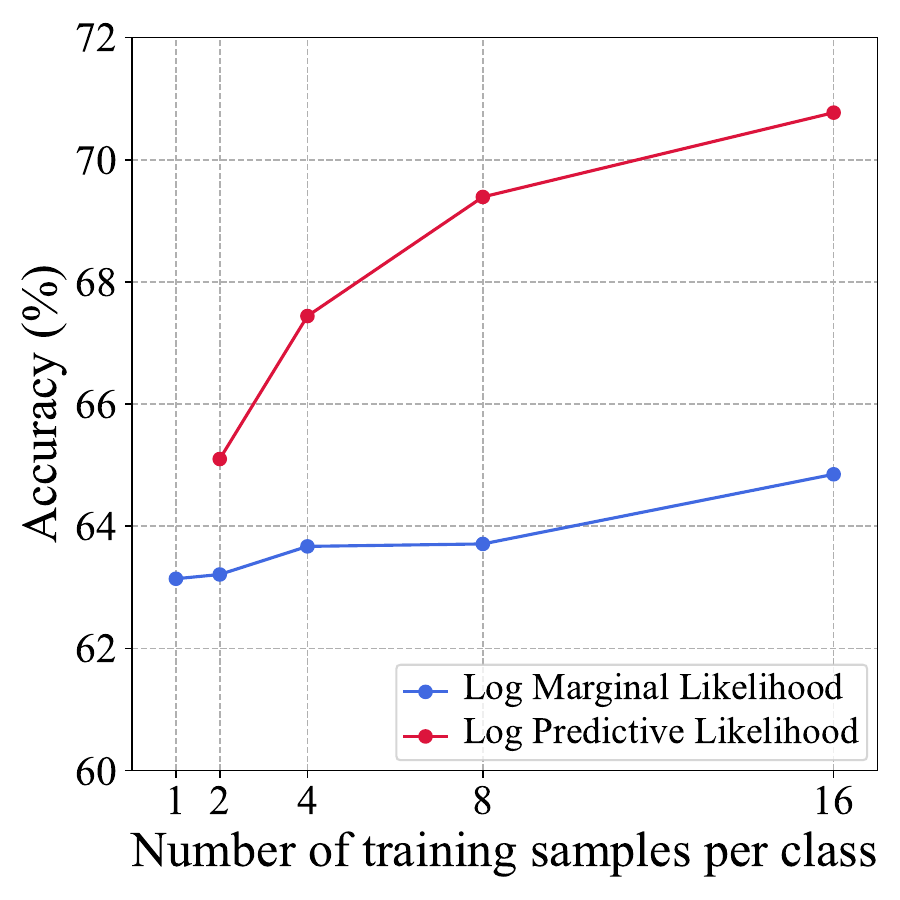}
        \caption{}
      \label{ablationobj}
    \end{subfigure}%
    \caption{Ablation studies on (a) GP mean, (b) GP base kernel, (c) Pre-trained model, and (d) hyper-parameter optimization objective. All experiments are conducted on ImageNet.}
    \vspace{-6pt}
    \label{fig:ablation}
\end{figure*}

\subsection{Model Calibration}
We then evaluate the model calibration of the proposed method by the ECE metric~\cite{guo2017calibration}.
For our method, we take the maximum element in $\mathbb{E}[\vf^*]$ as predictive confidence to calculate ECE.
The results are presented in \cref{tab:ECE}, and our model is slightly worse than Ens-CaFo. 
However, according to \cite{nixon2019measuring}, 
ECE leaves ambiguity in both its binning implementation and the calibration computation for multi-class scenarios.
Its robust variant, TACE~\cite{nixon2019measuring}, can be a better alternative. 
As shown in \cref{tab:ECE}, our method enjoys the best TACE compared to all baselines. 

\begin{table}[t]
    \setlength{\tabcolsep}{5pt}
    \centering
    \begin{tabular}{ccccc}
    \toprule
     Method & Ens-LP & Ens-LP$^\dagger$ & Ens-CaFo & Ours\\
    \midrule
    ECE & 0.1489& 0.1858 & \textbf{0.0577} & 0.0786 \\
    TACE & 0.0462 & 0.0477 & 0.0545 & \textbf{0.0169} \\
    \bottomrule
      \end{tabular}
     \captionof{table}{ECE and TACE of the four ensemble methods. All experiments are conducted on ImageNet.}
     \vspace{-5pt}
     \label{tab:ECE}
  \end{table}

We further present the reliability diagrams of the four methods in \cref{fig:ece}. 
The model calibration is good if the reliability diagram is close to the diagonal. 
As shown, compared to Ens-LP and Ens-LP$^\dagger$, our method and Ens-CaFo are more well calibrated. 
Besides, our method, Ens-LP, and Ens-LP$^\dagger$ tend to be underconfident, and the Ens-CaFo tends to be overconfident. 
Combining the results in \cref{tab:ECE} and \cref{fig:ece} yields the conclusion that our method enjoys good model calibration.

\vspace{-1pt}
\subsection{Ablation Study}
\label{sec:ablk}
In this section, we offer ablation studies for our method, including an examination of the mean and kernel of the GP, an investigation into how optimization objectives impact the results, and some visualization results.

\textbf{GP mean.}
We investigate the impact of the GP mean on the final results in \cref{ablationmean}. 
As shown, 
when the mean function equals zero or a learnable vector, prior knowledge is not incorporated into the GP model, and the final few-shot classification performance is unsatisfactory. 

\textbf{GP base kernel.}
We then delve into the specification of the GP base kernel.
The GP base kernel takes the RBF formula, and we opt for it due to its common usage, ease of implementation, and effectiveness. We also explore other formulas of base kernels, e.g., the Laplacian kernel and Mat\'ern kernel. 
The results on ImageNet are presented in \cref{ablationcov}.
It is evident that the RBF kernel performs best. 

\textbf{Pre-trained model.} To delve deeper into the impact of different pre-trained models, we test using various pre-trained models to specify the GP kernel. The results on ImageNet are shown in \cref{ablationkernel}. It is evident that integrating multiple sources of prior knowledge provided by different pre-trained models
leads to substantial advantages. We observe that assembling three pre-trained models already provides comprehensive prior knowledge of ImageNet, and additional integration of pre-trained models like SimCLR~\cite{chen2020simple} does not significantly improve performance. 
Including models pre-trained on datasets distinct from ImageNet can probably bring further benefits. 

\textbf{Objective.}
As pointed out, the marginal likelihood tends to be sensitive to prior assumptions, potentially resulting in underfitting or overfitting~\cite{lotfi2022bayesian,ke2023revisiting}. 
Therefore, the marginal likelihood can be negatively correlated with the generalization capability. 
Thus, we advocate the predictive likelihood for tuning hyper-parameters.
We perform an empirical study on the objective for hyper-parameter optimization in \cref{ablationobj}. 
As previously explained, under the 1-shot setting, we cannot use predictive likelihood.
The results clearly echo such an argument and support the use of predictive likelihood for hyper-parameter optimization.

\begin{figure}[t]
 \centering
    \begin{subfigure}{0.5\linewidth}
      \includegraphics[width=\linewidth]{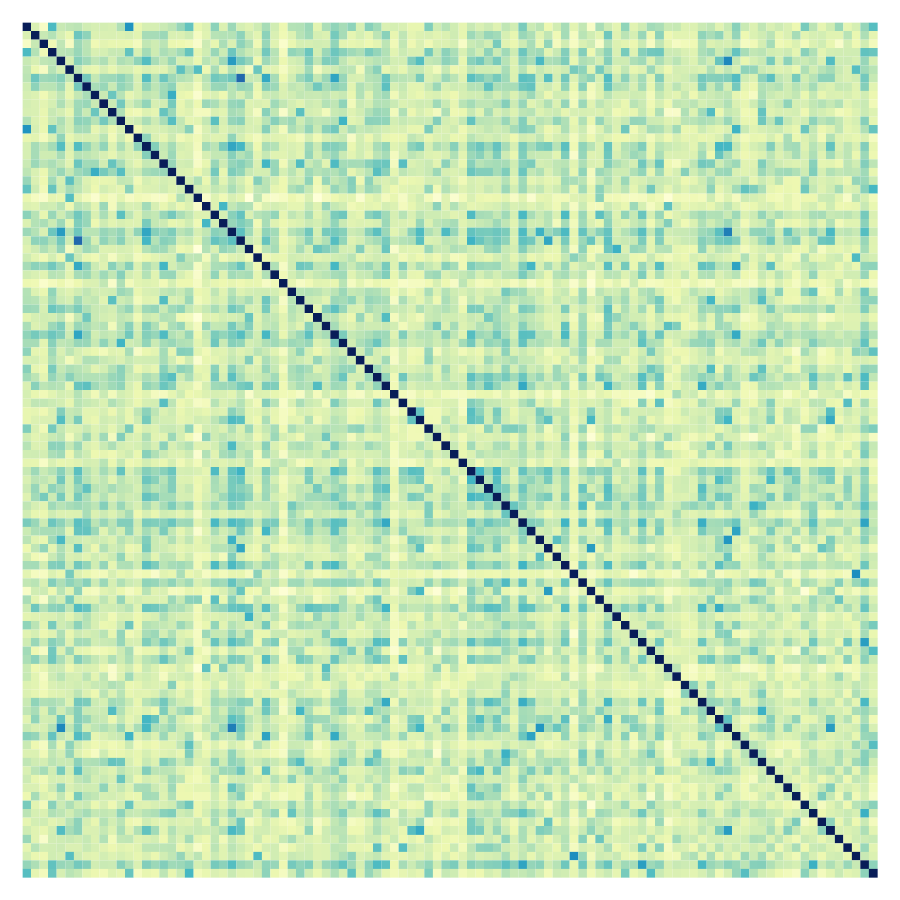}
      \caption{CLIP}
    \end{subfigure}%
    \begin{subfigure}{0.5\linewidth}
      \includegraphics[width=\linewidth]{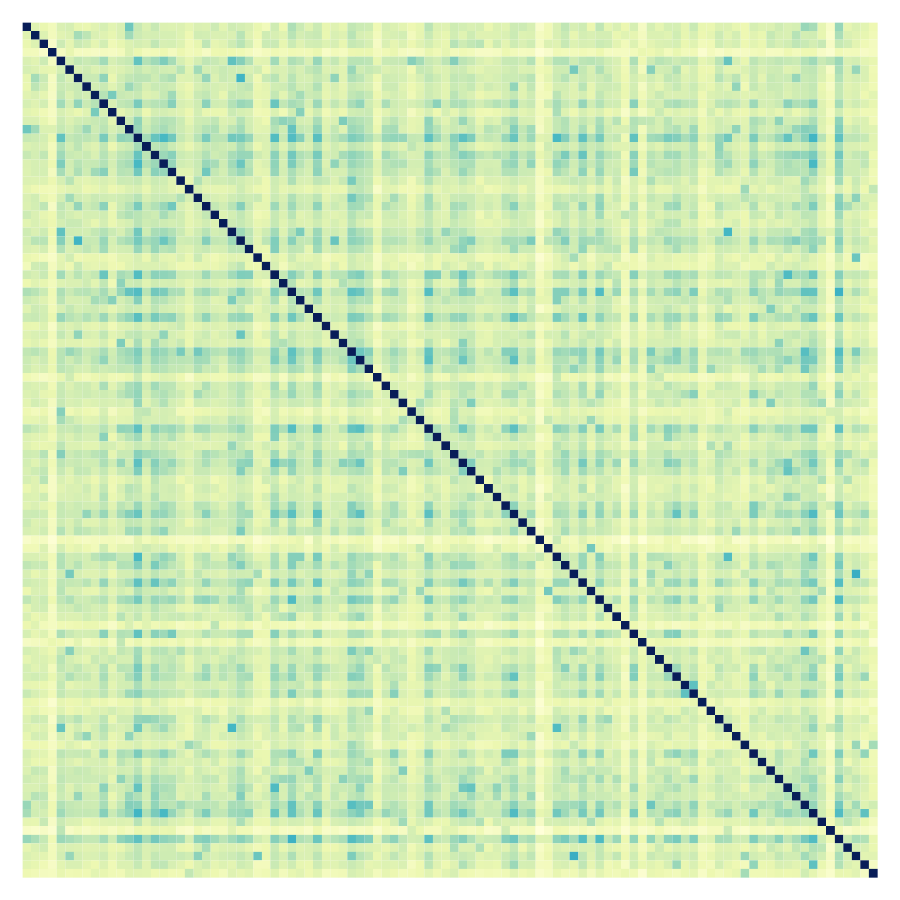}
      \caption{DINO}
    \end{subfigure} \\
    \begin{subfigure}{0.5\linewidth}
      \includegraphics[width=\linewidth]{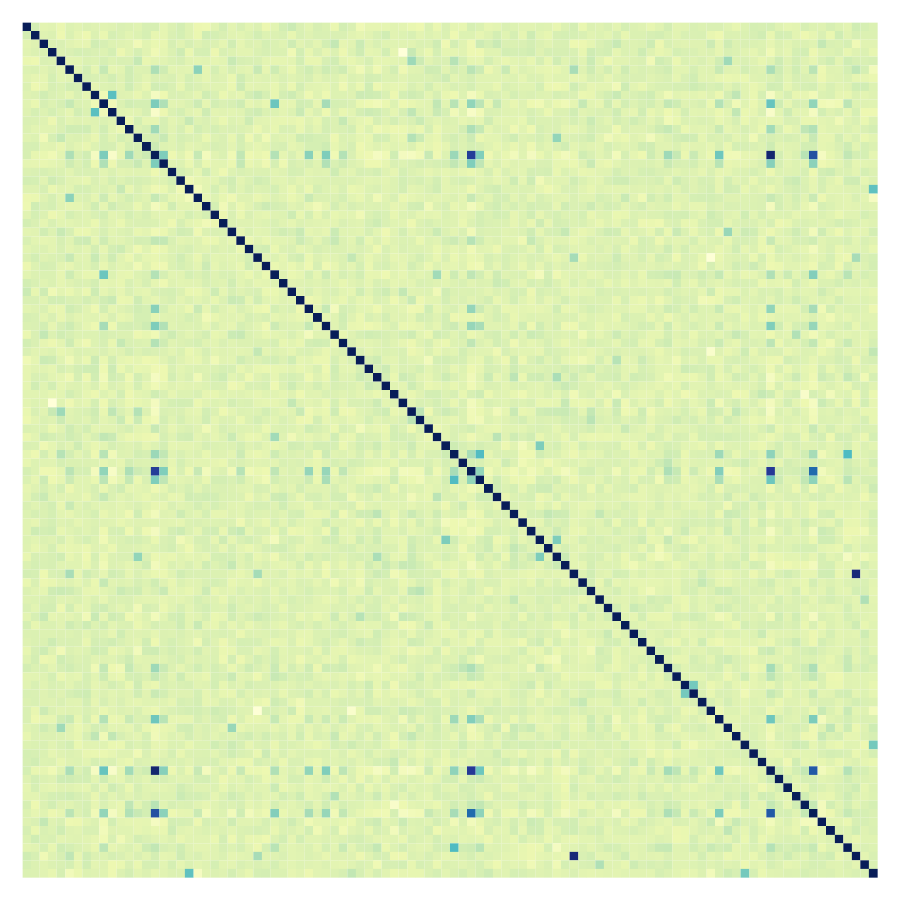}
      \caption{MoCo}
    \end{subfigure}%
    \begin{subfigure}{0.5\linewidth}
      \includegraphics[width=\linewidth]{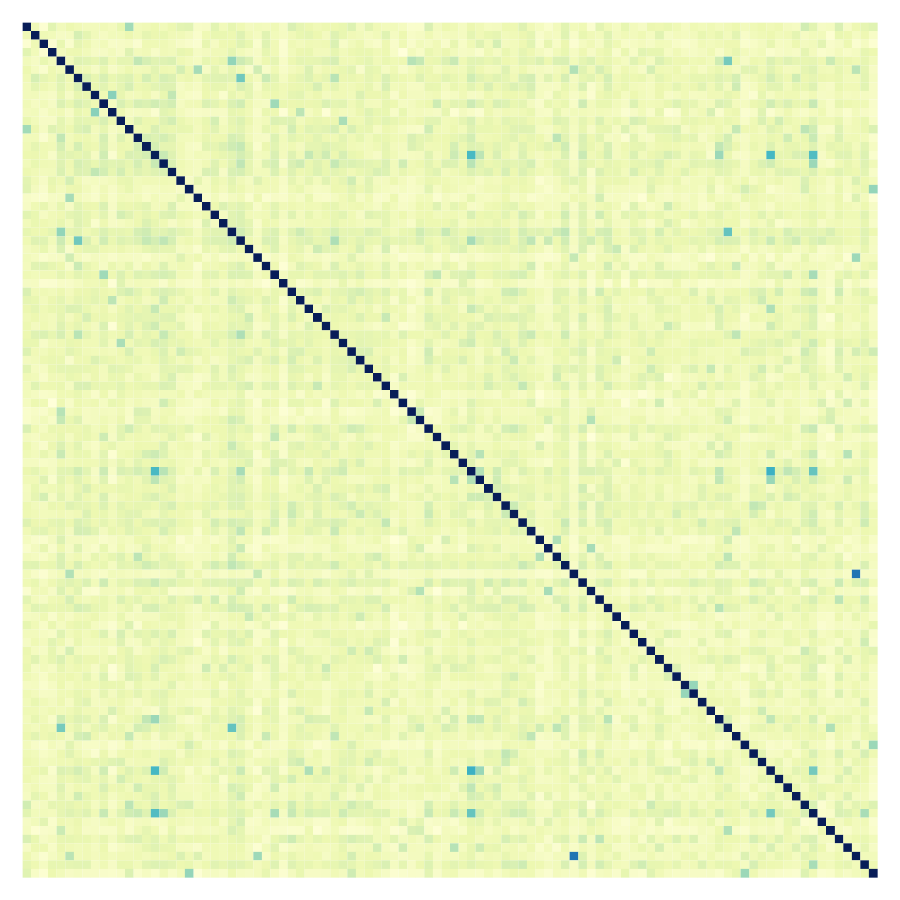}
      \caption{Ensemble}
    \end{subfigure}
    \caption{Visualization of prior kernel similarities.}
    \label{fig:prior}
    \vspace{-5pt}
\end{figure}

\textbf{Visualization of the prior kernels.}
In \cref{fig:prior}, we illustrate the data similarities given by the prior deep kernels defined with various pre-trained models.
The data points are randomly sampled from ImageNet.
The results reflect that distinct prior knowledge regarding data similarities is embedded in these models,
and through the kernel ensemble approach, our method can achieve knowledge integration.
\vspace{-2pt}
\section{Conclusion}
This work presents a simple and effective Bayesian approach for low-shot image classification. 
We develop a GP framework to flexibly incorporate diverse prior knowledge from pre-trained models. 
Extensive experiments showcase the superiority and strong generalization capabilities of our method. 
More importantly, we demonstrate that the uncertainty given by our method is well-calibrated. Our method will likely enable intriguing applications such as OOD detection by leveraging the uncertainty estimates.
Overall, our study demonstrates the exceptional power of Bayesian methods in the large model era and aids in paving the path for future algorithmic improvements in low-shot learning.

\section*{Acknowledgments}
This work was supported by NSF of China (No. 62306176, 62106121), Natural Science Foundation of Shanghai (No. 23ZR1428700), the Key Research and Development Program of Shandong Province, China (No. 2023CXGC010112), CCF-Baichuan-Ebtech Foundation Model Fund, and the MOE Project of Key Research Institute of Humanities and Social Sciences (22JJD110001). 

{   
    \small
    \bibliographystyle{ieeenat_fullname}
    \bibliography{main}
}

\clearpage
\appendix

\section{Datasets Preparation}
The datasets employed in this work have been slightly modified to accommodate low-shot classification better. To ensure a fair comparison with previous works, in line with CaFo~\cite{zhang2023prompt}, we randomly sampled 1, 2, 4, 8, and 16 data points per class from ImageNet~\cite{deng2009imagenet}. These sets are designated as 1, 2, 4, 8, and 16-shot training sets, with the ImageNet validation set serving as the test set.
All samples from ImageNet-V2~\cite{recht2019imagenet} and ImageNet-Sketch~\cite{hendrycks2021natural} are exclusively used for testing purposes.
For other datasets, we adhere to the same train/test/val splits as established by CaFo.

\section{Additional Ablation Study}
\textbf{CLIP’s Visual Encoders. }
For further performance enhancement on ImageNet~\cite{deng2009imagenet}, we attempt to change the backbone of the image encoder in CLIP from ResNet-50 to ViT-B/16. 
We provide the corresponding results in \cref{tab:vit}. 
It is easy to see that our method remains to surpass all the ensemble baselines consistently.

\begin{table}[h]
  \centering
  \begin{tabular}{lccccc}
    \toprule
    \multicolumn{1}{l}{Shot} & 1 & 2 & 4 & 8 & 16              \\
    \midrule
    Ens-LP    &41.60 & 51.75 & 59.82& 65.42&69.86    \\
    Ens-LP$\dagger$    &69.81&	71.11&	71.45&	73.05&	74.20  \\
    Ens-CaFo    &70.00	&71.03&	71.79	&72.86&	74.49   \\
    Ours   & \textbf{70.70}  & \textbf{71.48}  & \textbf{72.62}  & \textbf{73.96}  & \textbf{75.22}  \\
    \bottomrule
  \end{tabular}
  \caption{Accuracy (\%) on ImageNet when using the CLIP with a ViT-B/16 image encoder.}
    \label{tab:vit}
\end{table}

\textbf{DALL-E Augmentation. }
Following CaFo~\cite{zhang2023prompt}, we also explore the impact of using synthetic images for data augmentation. According to ~\cite{zhang2023prompt}, under the 1,2,4-shot setting, we use 8 synthetic images per class for augmentation. Under the 8, 16-shot setting, we use 2 synthetic images per class for augmentation. The results in \cref{tab:dalle} can serve as an ablation study on the DALL-E~\cite{ramesh2021zero} augmentation. We can see that the use of synthetic images is intended to provide benefits when dealing with an extremely limited number of training samples, e.g., 1 or 2-shot setting. 
With data augmentation, our method also consistently outperforms other baselines. This demonstrates the effectiveness of our approach as well as its robustness against data augmentation.

\begin{table}[h]
  \centering
  \begin{tabular}{lccccc}
    \toprule
    \multicolumn{1}{l}{Shot} & 1 & 2 & 4 & 8 & 16              \\
    \midrule
    Ens-LP    &56.23&	57.70& 59.60 &63.67	&67.23    \\
    Ens-LP$\dagger$    &66.62&	67.08&	67.20&	67.71&	69.22  \\
    Ens-CaFo    &65.19	&66.02&	66.65	&67.45&	68.85   \\
    Ours   & \textbf{67.32}  & \textbf{67.93}  & \textbf{68.65}  & \textbf{69.56}  & \textbf{70.83}  \\
    \bottomrule
  \end{tabular}
  \caption{Accuracy (\%) on ImageNet when using DALL-E augmentation.}
    \label{tab:dalle}
\end{table}

\section{Visualization of Uncertainty Estimates}
\begin{figure}[h]
  \centering
    \includegraphics[width={\linewidth}]{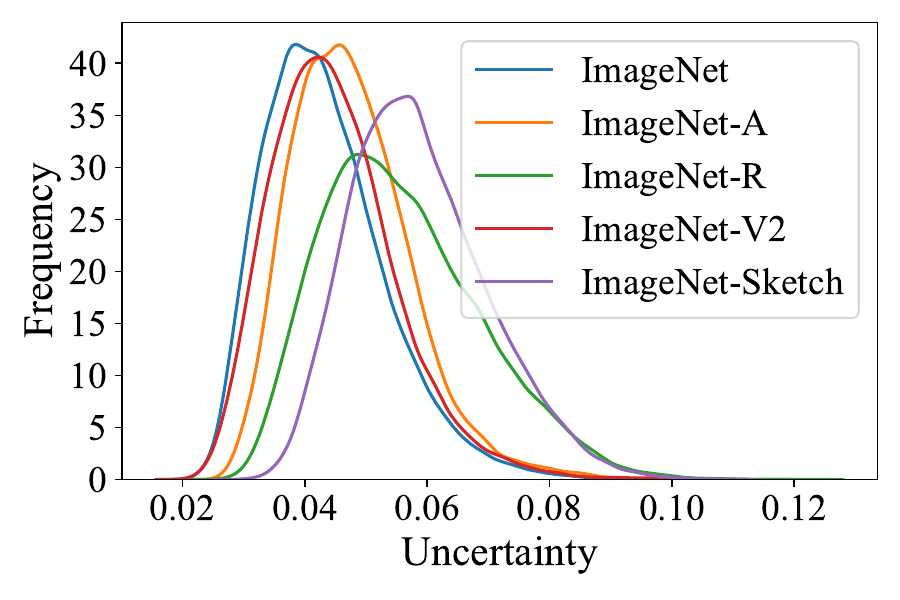}
   \caption{Histogram for uncertainty estimates. We evaluate our methods on ImageNet, ImageNet-V2, ImageNet-A, ImageNet-R, and Imagenet-Sketch.}
   \label{fig:our_5uncs}
\end{figure}

We train our model on ImageNet~\cite{deng2009imagenet} and then test on ImageNet-V2~\cite{recht2019imagenet}, ImageNet-A~\cite{hendrycks2021natural}, ImageNet-R~\cite{hendrycks2021many}, and Imagenet-Sketch~\cite{hendrycks2021natural} to get the uncertainty estimate distributions. The results in \cref{fig:our_5uncs} align with the fact that ImageNet-V2 and ImageNet-A have similar distributions with ImageNet, while the distributions of ImageNet-R and Imagenet-Sketch are different from ImageNet.

\begin{figure*}[t]
 \centering
    \begin{subfigure}{0.33\linewidth}
      \includegraphics[width=\linewidth]{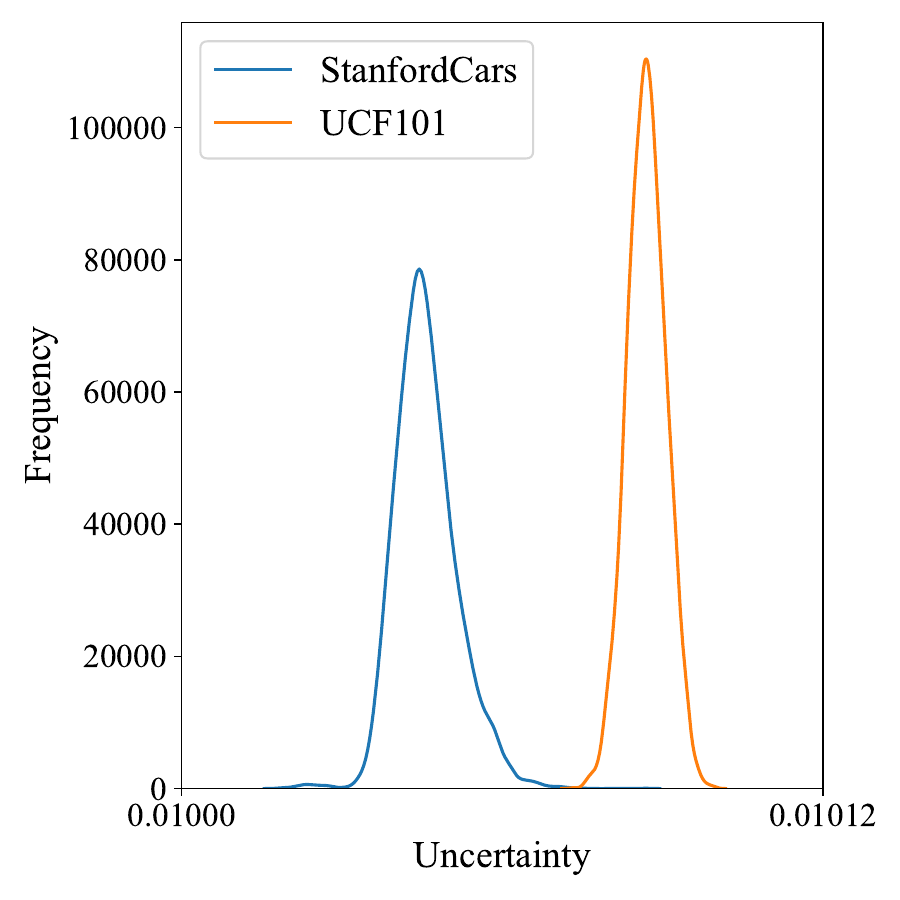}
    \end{subfigure}%
    \begin{subfigure}{0.33\linewidth}
      \includegraphics[width=\linewidth]{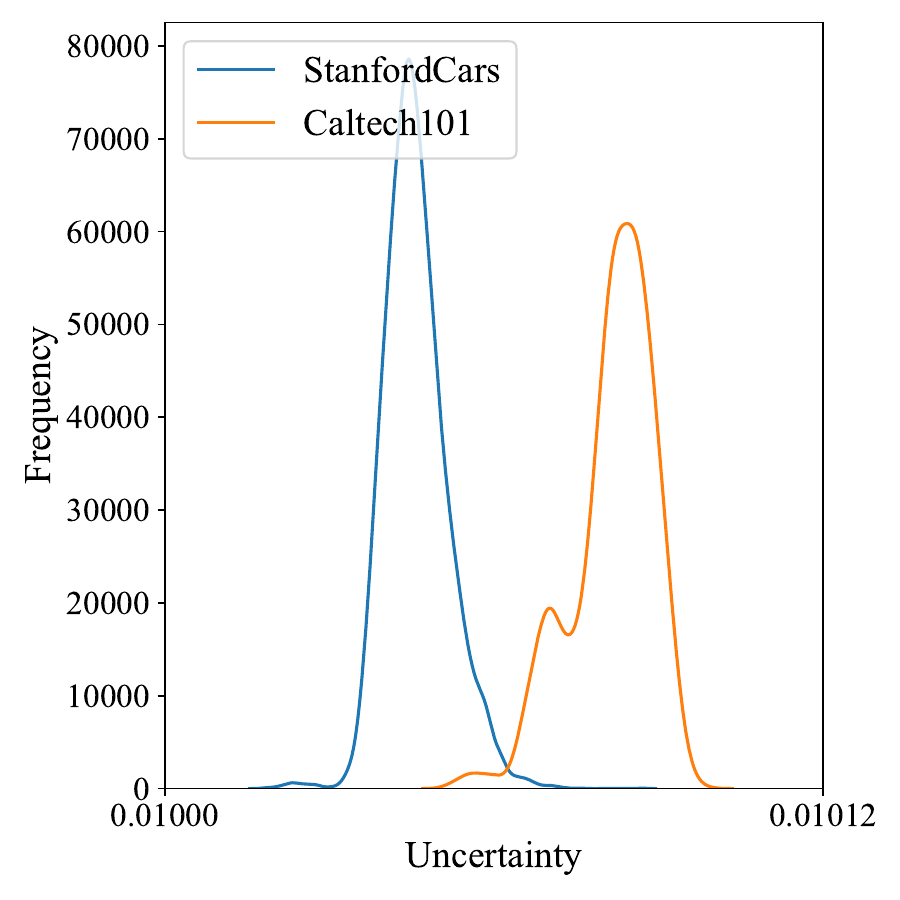}
    \end{subfigure}%
    \begin{subfigure}{0.33\linewidth}
      \includegraphics[width=\linewidth]{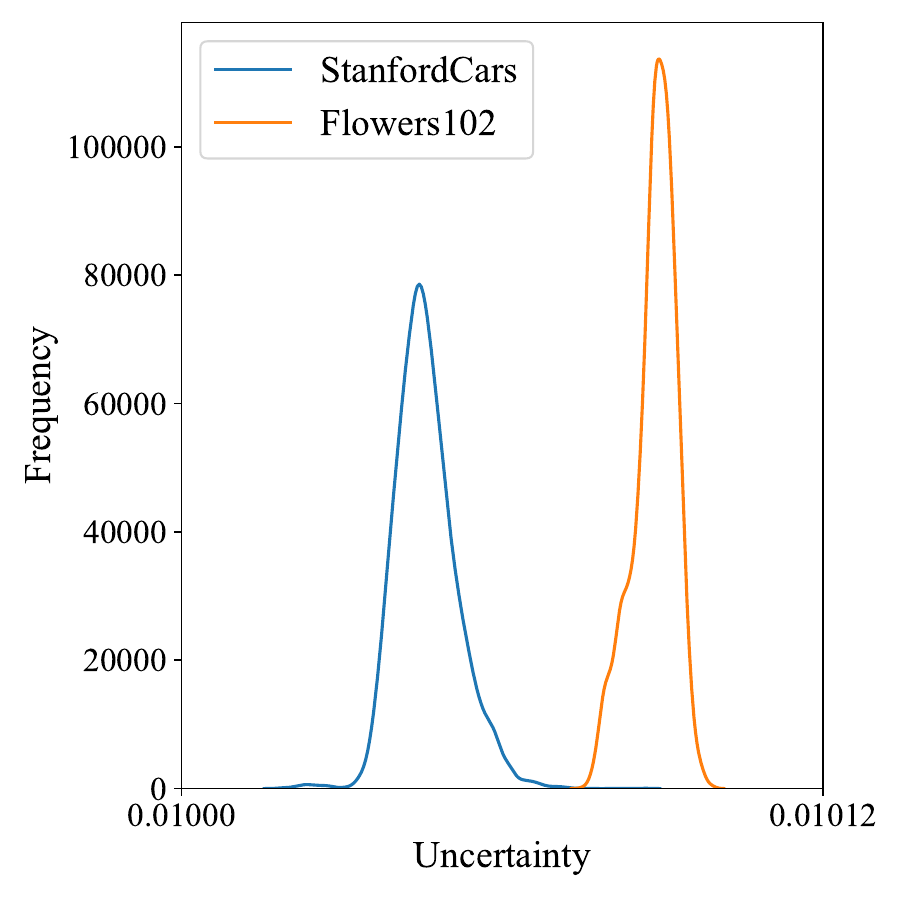}
    \end{subfigure} \\
    \begin{subfigure}{0.33\linewidth}
      \includegraphics[width=\linewidth]{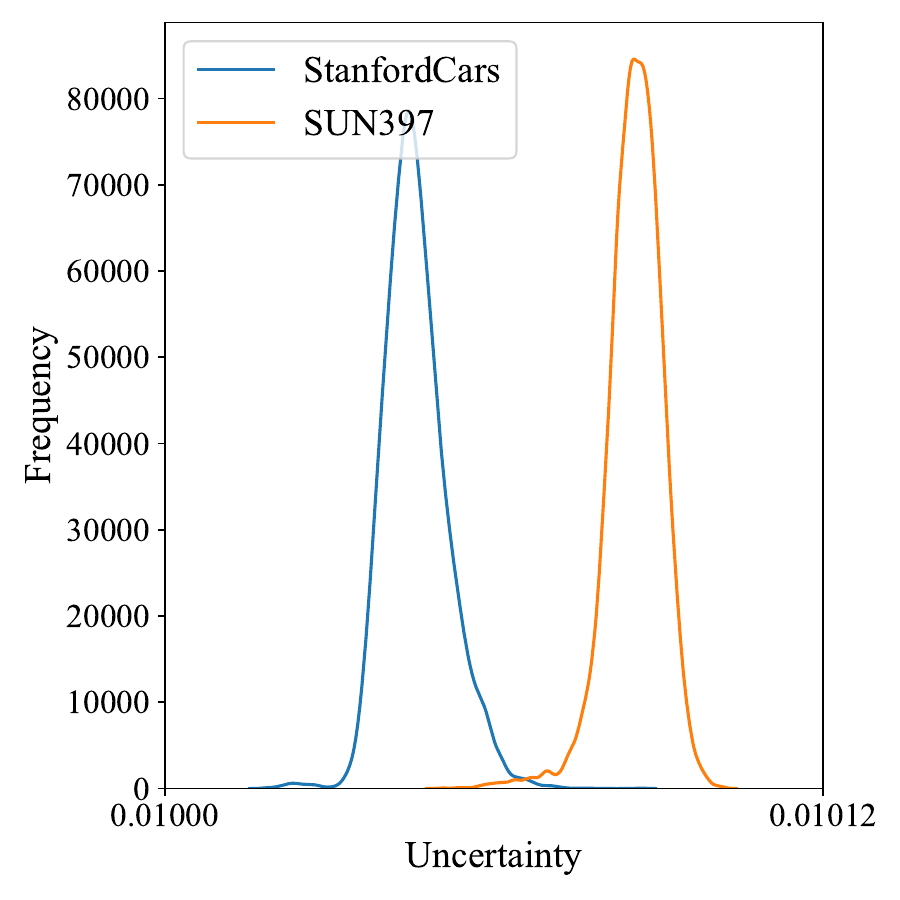}
    \end{subfigure} 
    \begin{subfigure}{0.33\linewidth}
      \includegraphics[width=\linewidth]{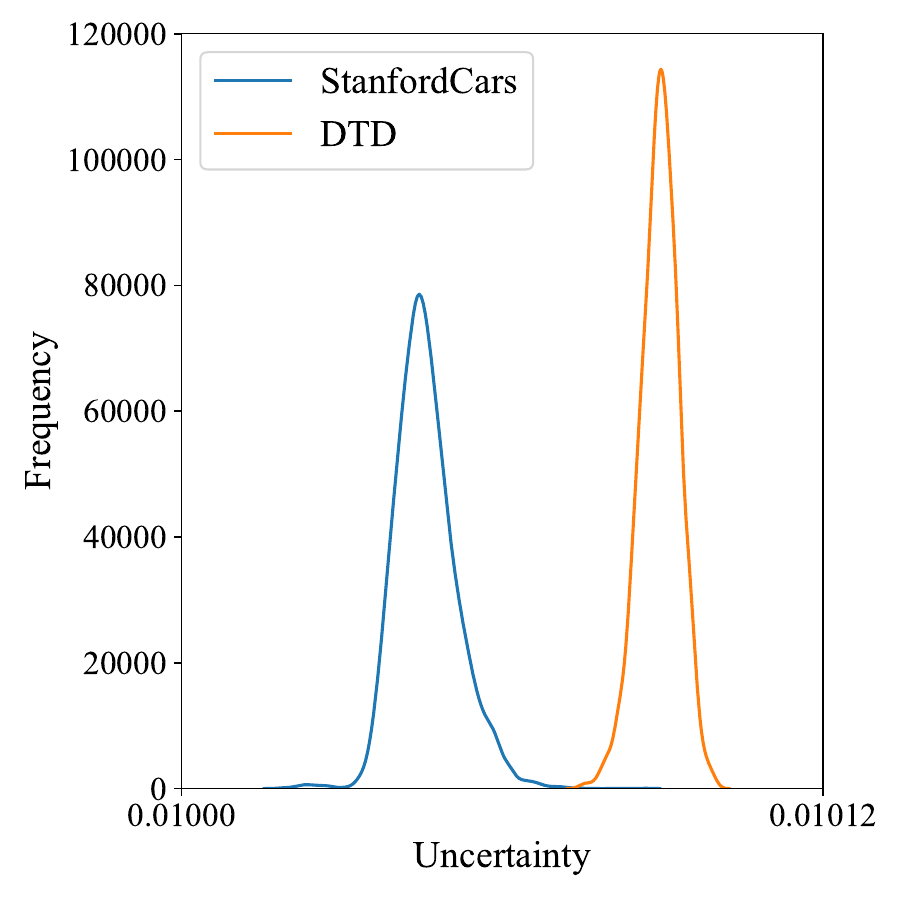}
    \end{subfigure}%
    \begin{subfigure}{0.33\linewidth}
      \includegraphics[width=\linewidth]{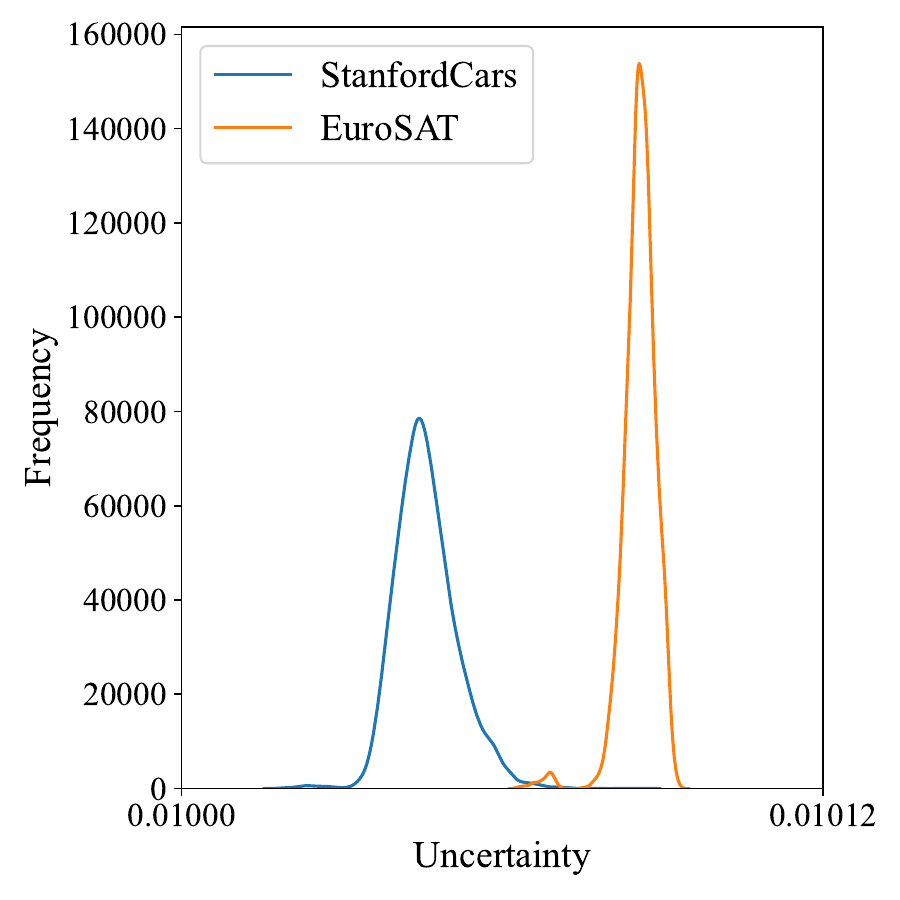}
    \end{subfigure} \\
    \begin{subfigure}{0.33\linewidth}
      \includegraphics[width=\linewidth]{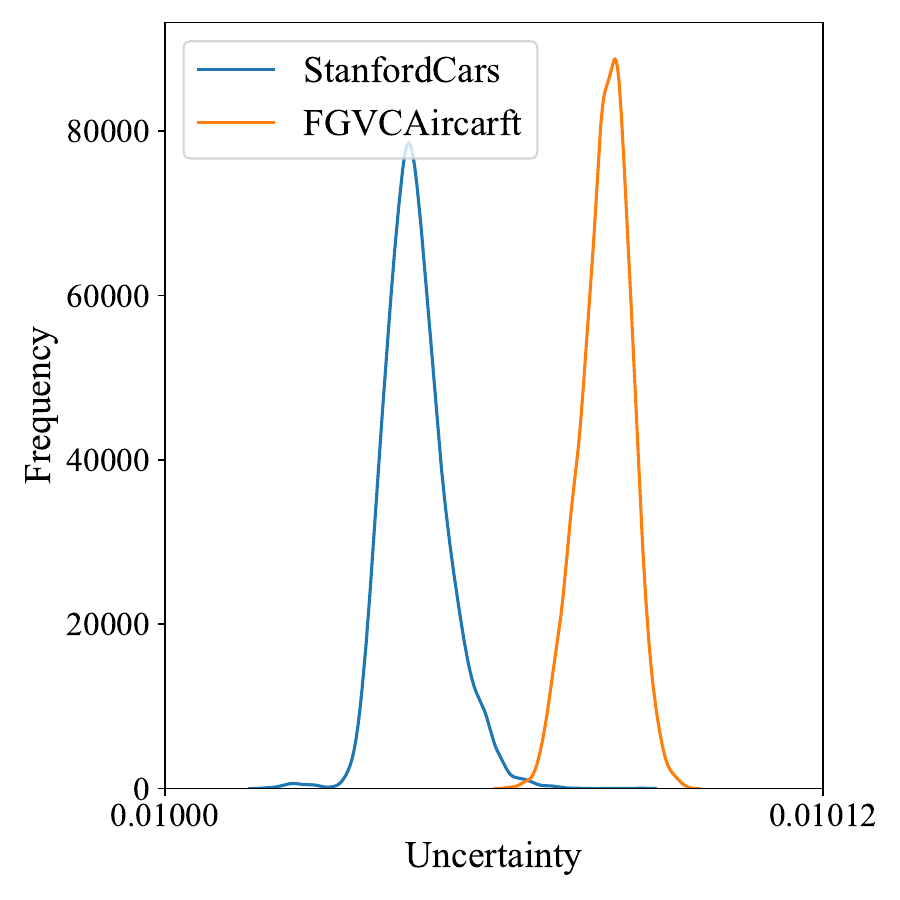}
    \end{subfigure}%
    \begin{subfigure}{0.33\linewidth}
      \includegraphics[width=\linewidth]{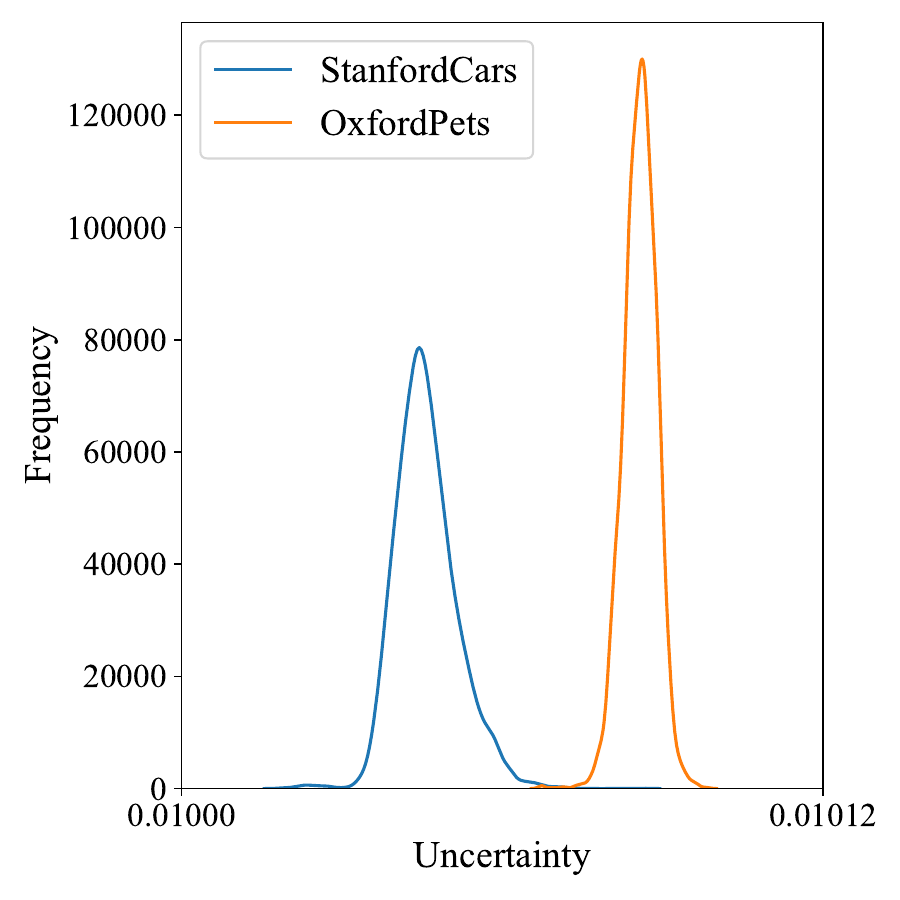}
    \end{subfigure}%
    \begin{subfigure}{0.33\linewidth}
      \includegraphics[width=\linewidth]{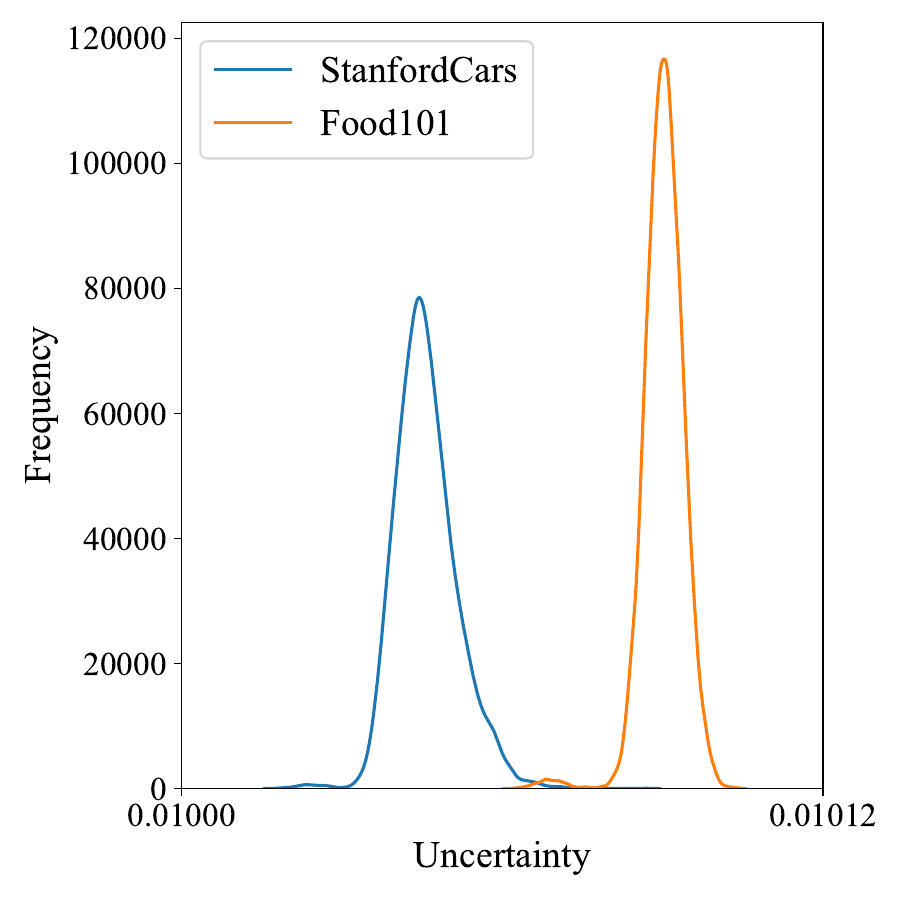}
    \end{subfigure}%
    \caption{Histogram for uncertainty estimates. We evaluate our
    methods on StanfordCars and nine other datasets.}
    \label{fig:uncs_cars}
\end{figure*}

To further evaluate the OOD detection capability of our method,  
we initially pre-train our model using the StanfordCars~\cite{krause20133d} dataset and subsequently evaluate its performance on various datasets to get histograms for uncertainty estimates. As depicted in \cref{fig:uncs_cars}, it is evident that our model distinguishes unique uncertainty distributions among the nine datasets and the StanfordCars dataset. The findings suggest that our model discerns dissimilarities, classifying the nine datasets as OOD data from the StanfordCars dataset.

\end{document}